\documentclass[acmtog]{acmart}

\usepackage{booktabs}

\usepackage[ruled]{algorithm2e} 

\SetAlFnt{\small}
\SetAlCapFnt{\small}
\SetAlCapNameFnt{\small}
\SetAlCapHSkip{0pt}

\usepackage{xcolor}
\definecolor{myred}{rgb}{0.8,0,0}
\definecolor{mygreen}{rgb}{0,0.8,0}
\usepackage{pifont}
\newcommand{\cmark}{{\color{mygreen}\ding{51}}}
\newcommand{\xmark}{{\color{myred}\ding{55}}}

\acmJournal{TOG}
\acmArticle{94}
\acmYear{2020}
\acmVolume{40}
\acmNumber{4}
\acmMonth{7} 
\acmDOI{10.1145/3450626.3459749}
\acmISBN{978-1-4503-8374-5/21/08}

\setcopyright{rightsretained} 

\begin{document}


\title{Real-time Deep Dynamic Characters}


\author{Marc Habermann, Lingjie Liu}
\affiliation{%
	\institution{Max Planck Institute for Informatics}
	\streetaddress{Campus E1, Stuhlsatzenhausweg 4}
	\city{Saarbruecken}
	\state{Saarland, Germany}
	\postcode{66123}
}
\email{mhaberma@mpi-inf.mpg.de}
\email{lliu@mpi-inf.mpg.de}

\author{Weipeng Xu, Michael Zollhoefer}
\affiliation{%
	\institution{Facebook Reality Labs}
	\streetaddress{District Fifteen: 131 15th Street}
	\city{Pittsburgh}
	\state{Pennsylvania, USA}
	\postcode{15222}
}
\email{wxu@mpi-inf.mpg.de}
\email{zollhoefer@fb.de}

\author{Gerard Pons-Moll, and Christian Theobalt}
\affiliation{%
	\institution{Max Planck Institute for Informatics}
	\streetaddress{Campus E1, Stuhlsatzenhausweg 4}
	\city{Saarbruecken}
	\state{Saarland, Germany}
	\postcode{66123}
}
\email{gpons@mpi-inf.mpg.de}
\email{theobalt@mpi-inf.mpg.de}


\renewcommand{\shortauthors}{Habermann et al.}


%
\begin{abstract}
We propose a deep videorealistic 3D human character model displaying highly realistic shape, motion, and dynamic appearance learned in a new weakly supervised way from multi-view imagery. 
In contrast to previous work, our controllable 3D character displays dynamics, e.g., the swing of the skirt, dependent on skeletal body motion in an efficient data-driven way, without requiring complex physics simulation. 
Our character model also features a learned dynamic texture model that accounts for photo-realistic motion-dependent appearance details, as well as view-dependent lighting effects.
During training, we do not need to resort to difficult dynamic 3D capture of the human; instead we can train our model entirely from multi-view video in a weakly supervised manner.
To this end, we propose a parametric and differentiable character representation which allows us to model coarse and fine dynamic deformations, e.g., garment wrinkles, as explicit space-time coherent mesh geometry that is augmented with high-quality dynamic textures dependent on motion and view point.
As input to the model, only an arbitrary 3D skeleton motion is required, making it directly compatible with the established 3D animation pipeline. 
We use a novel graph convolutional network architecture to enable motion-dependent deformation learning of body and clothing, including dynamics, and a neural generative dynamic texture model creates corresponding dynamic texture maps.
We show that by merely providing new skeletal motions, our model creates motion-dependent surface deformations, physically plausible dynamic clothing deformations, as well as video-realistic surface textures at a much higher level of detail than previous state of the art approaches, and even in real-time.
\end{abstract}
%



\copyrightyear{2021} 




\maketitle


\begin{figure}
	\includegraphics[width=0.90\linewidth]{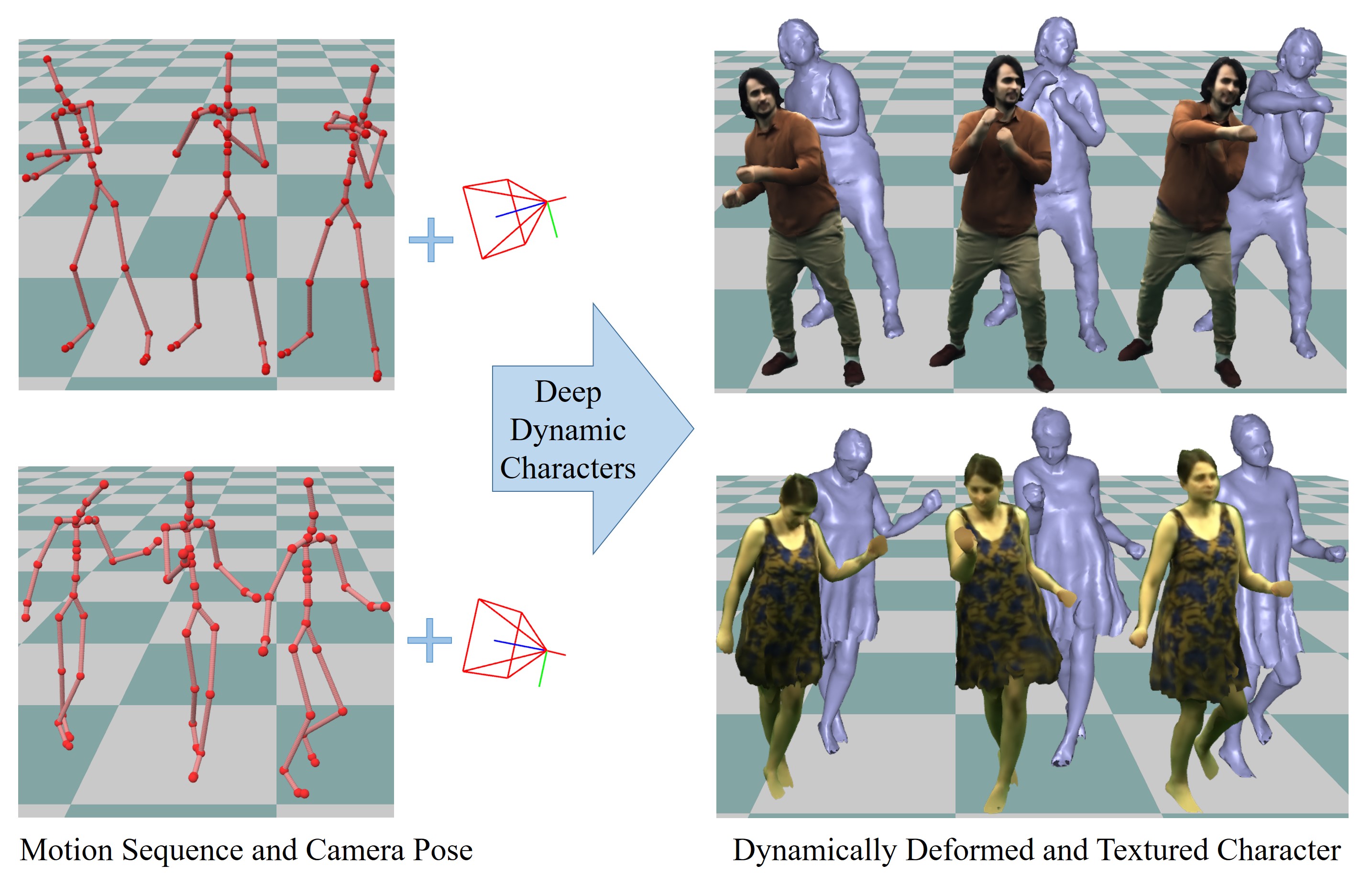} 
	\caption
	{
		Our learning-based method takes a sequence of poses and regresses the motion- and view-dependent dynamic surface deformation and texture of a person-specific template which looks video realistic.
	}
	\label{fig:teaser}
	\vspace{-18pt}
\end{figure}


%
\section{Introduction}
\label{sec:introduction}
%
%
%
Animatable and photo-realistic virtual 3D characters are of enormous importance nowadays. 
With the rise of computer graphics in movies, games, telepresence, and many other areas, 3D virtual characters are everywhere. 
Recent developments in virtual and augmented reality and the resulting immersive experience further boosted the need for virtual characters as they now become part of our real lives. 
However, generating realistic characters still requires manual intervention, expensive equipment, and the resulting characters are either difficult to control or not realistic.
Therefore, our goal is to learn digital characters which are both realistic and easy to control and can be learned directly from a multi-view video. 
%
%
%
\par
It is a complicated process to synthesize realistic looking images of deforming characters following the conventional computer graphics pipeline.
The static geometry of real humans is typically represented with a mesh obtained with 3D scanners. 
In order to pose or animate the mesh, a skeleton has to be attached to the geometry, i.e. rigging, and skinning techniques can then be used to deform the mesh according to the skeletal motion.
While these approaches are easy to control and efficient, they lack realism as the non-rigid deformations of clothing are not modeled, e.g., the swinging of a skirt.
While physics simulation can address this, it requires expert knowledge as it is hard to control.
Further, these techniques are either computationally expensive or not robust to very articulated poses leading to glitches in the geometry.
Finally, expensive physically based rendering techniques are needed to render realistic images of the 3D character.
Those techniques are not only time consuming, but also require expert knowledge and manual parameter tuning.
%
%
%
\par
To model clothing deformations, recent work combines classical skinning with a learnt mapping from skeletal motion to non-rigid deformations, and learns the model from data.
One line of work learns from real data, but results either lack realism~\cite{ma20autoenclother} or are limited to partial clothing, e.g., a T-shirt~\cite{lhner18}.
More importantly, as they rely on ground truth registered 3D geometry, they require expensive 3D scanners and challenging template registration.
Another line of work tries to learn from a large database of simulated clothing~\cite{patel20,guan12}.
While they can generalize across clothing categories and achieve faster run-times than physics simulations, the realism is still limited by the physics engine used for training data generation. 
%
%
%
\par
Furthermore, the texture dynamics are not captured by the aforementioned methods, although they are crucial to achieve photo-realism.
Monocular neural rendering approaches \cite{liu19,liu20, chan2019dance} for humans learn a mapping from a CG rendering to a photo-realistic image, but their results have limited resolution and quality and struggle with consistency when changing pose and viewpoint.
The most related works~\cite{Xu:SIGGRPAH:2011,casas14,shysheya19} are the ones leveraging multi-view imagery for creating animatable characters. 
However, all of them are not modeling a motion-dependent deforming geometry and/or view-dependent appearance changes.
%
%
%
\par
To overcome the limitations of traditional skinning, the requirement of direct 3D supervision of recent learning based methods, as well as their lack of dynamic textures, we propose a learning based method that predicts the non-rigid character surface deformation of the full human body as well as a dynamic texture from skeletal motion \emph{using only weak supervision in the form of multi-view images during training}.
At the core of our method is a differentiable character (with neural networks parameterizing dynamic textures and non-rigid deformations) which can generate images differentiable with respect to its parameters.
This allows us to train directly with multi-view image supervision using analysis by synthesis and back-propagation, instead of pre-computing 3D mesh registrations, which is difficult, tedious and prone to error. 
In designing differentiable characters, our key insight is to learn as much of the deformation as possible in geometry space, and produce the subtle fine details in texture space. 
Compared to learning geometry deformations in image space, this results in much more coherent results when changing viewpoint and pose. 
To this end, we propose a novel graph convolutional network architecture which takes a temporal motion encoding and predicts the surface deformation in a coarse to fine manner using our new fully differentiable character representation.
The learned non-rigid deformation and dynamic texture not only account for dynamic clothing effects such as the swinging of a skirt caused by the actor's motion, or fine wrinkles appearing in the texture, but also fixes traditional skinning artifacts such as candy wrappers.
Moreover, as our dynamic texture is conditioned on the camera pose, our approach can also model view-dependent effects, e.g., specular surface reflections.
In summary, our contributions are:
\begin{itemize}
	\item{The first learning based real-time approach that takes a motion and camera pose as input and predicts the motion-dependent surface deformation and motion- and view-dependent texture for the full human body using direct image supervision.}
	\item{A differentiable 3D character representation which can be trained from coarse to fine (Sec.~\ref{sec:characterModel}).}
	\item{A graph convolutional architecture allowing to formulate the learning problem as a graph-to-graph translation (Sec.~\ref{sec:egnet}).}
	\item{We collected a new benchmark dataset, called \textit{DynaCap}, containing 5 actors captured with a dense multi-view system which we will make publicly available (Sec.~\ref{sec:datasets}).}
\end{itemize}
Our dynamic characters can be driven either by motion capture approaches or by interactive editing of the underlying skeleton.
This enables many exciting applications in gaming and movies, such as more realistic character control as the character deformation and texture will account for dynamic effects.
Our qualitative and quantitative results show that our approach is clearly a step forward towards photo-realistic and animatable full body human avatars.

\section{Related Work}
%
%
%
%
While some works focus on adapting character motions to a new geometric environment, e.g. walking on a rough terrain,~\cite{10.1145/3072959.3073663} or  create character motions from goal oriented user instructions~\cite{ Starke2019NeuralSM}, we assume the motion is given and review works that focus on animatable characters, learning based cloth deformations, and differentiable rendering.
%
%
\paragraph{Video-based Characters}
Previous work in the field of video-based characters aims at creating photo-realistic renderings of controllable characters. 
Classical methods attempt to achieve this by synthesizing textures on surface meshes and/or employing image synthesis techniques in 2D space. 
Some works~\cite{carranza03,Li:2014,zitnick04,collet15,hilsmann20} focus on achieving free-viewpoint replay from multi-view videos with or without 3D proxies, however, they are not able to produce new poses for human characters. 
The approach of \cite{stoll10} incorporates a physically-based cloth model to reconstruct a rigged fully-animatable character in loose cloths from multi-view videos, but it can only synthesize a fixed static texture for different poses. 
To render the character with dynamic textures in new poses from arbitrary viewpoints, \cite{Xu:SIGGRPAH:2011} propose a method that  first retrieves the most similar poses and viewpoints in a pre-captured database and then applies retrieval based texture synthesis. 
However, their method takes several seconds per frame and thus cannot support interactive character animation. 
\cite{casas14,Volino2014} compute a temporally coherent layered representation of appearance in texture space to achieve interactive speed, but the synthesis quality is limited due to the coarse geometric proxy. 
Most of the traditional methods for free-viewpoint rendering of video-based characters fall either short in terms of generalization to new poses and/or suffer from a high runtime, and/or a limited synthesis quality. 
%
%
\par
More recent works employ neural networks to close the gap between rendered virtual characters and real captured images. 
While some approaches have shown convincing results for the facial area~\cite{lombardi18,kim2018deep}, creating photo-real images of the entire human is still a challenge.
Most of the methods, which target synthesizing entire humans, learn an image-to-image mapping from renderings of a skeleton~\cite{chan2019dance, Pumarola_2018_CVPR, Esser_2018_ECCV_Workshops, Si_2018_CVPR}, depth map~\cite{Martin18}, dense mesh~\cite{liu19,liu20,wang2018vid2vid,Sarkar2020} or joint position heatmaps~\cite{Lischinski2018}, to real images.
Among these approaches, the most related work~\cite{liu20} achieves better temporally-coherent dynamic textures by first learning fine scale details in texture space and then translating the rendered mesh with dynamic textures into realistic imagery. 
While only requiring a single camera, these methods only demonstrate the rendering from a fixed camera position while the proposed approach works well for arbitrary view points and also models the view-dependent appearance effects.
Further, these methods heavily rely on an image-to-image translation network to augment the realism, however, this refinement simply applied in 2D image space leads to missing limbs and other artifacts in their results.
In contrast, our method does not require any refinement in 2D image space but explicitly generates high-quality view- and motion-dependent geometry and texture for rendering to avoid such kind of artifacts. 
Similar to us, Textured Neural Avatars~\cite{shysheya19} (TNA) also assumes multi-view imagery is given during training. 
However, TNA is neither able to synthesize motion- and view-dependent dynamic textures nor to predict the dense 3D surface. 
Our method can predict motion-dependent deformations on surface geometry as well as dynamic textures from a given pose sequence and camera view leading to video-realistic renderings. 
%
%
\paragraph{Learning Based Cloth Deformation}
Synthesizing realistic cloth deformations with physics-based simulation has been extensively explored~\cite{Choi05, Nealen05, Narain12, Tang18, liang19, tao19, su20}. 
They employ either continuum mechanics principles followed by finite element discretization, or physically consistent models. 
However, they are computationally expensive and often require manual parameter tuning. 
To address this issue, some methods~\cite{kim08, wang10,fengww10, guan12, zurdo13, xuww14, Hahn14} model cloth deformations as a function of the underlying skeletal pose and/or the shape of the person and learn the function from data. 
%
%
\par
With the development of deep learning, skinning based deformations can be improved~\cite{Bailey:2018:FDD} over the traditional methods like linear blend skinning~\cite{Magnenat-Thalmann1988:4}.
Other works go beyond skinning based deformations and incorporate deep learning for predicting cloth deformations and learn garment deformations from the body pose and/or shape.
Some works~\cite{alldieck18a, alldieck18b, alldieck19, bhatnagar19,jin18,pons17} generate per-vertex displacements over a parametric human model to capture the garment deformations.
While this is an efficient representation, it only works well for tight cloths such as pants and shirts. 
\cite{gundogdu19} use neural networks to extract garment features at varying levels of detail (i.e., point-wise, patch-wise and global features). 
\cite{patel20} decompose the deformation into a high frequency and a low frequency component. 
While the low-frequency component is predicted from pose, shape and style of garment geometry with an MLP, the high-frequency component is generated with a mixture of shape-style specific pose models.
Related to that \citet{Choi_2020_ECCV_Pose2Mesh} predicts the geometry of the naked human from coarse to fine given the skeletal pose.
\citet{santesteban19} separate the global coarse garment fit, due to body shape, from local detailed garment wrinkles, due to both body pose dynamics and shape. 
Other methods~\cite{lhner18, zhangm20} recover fine garment wrinkles for high-quality renderings or 3D modeling by augmenting a low-resolution normal map of a garment with high-frequency details using GANs.
\citet{zhi2020texmesh} also reconstructs albedo textures and refines a coarse geometry obtained from RGB-D data.
Our method factors cloth deformation into low-frequency large deformations represented by an embedded graph and high-frequency fine wrinkles modeled by a per-vertex displacements, which allows for synthesizing deformations for any kind of clothing, including also loose clothes. 
In contrast to the above methods, our approach does not only predict geometric deformations but also a dynamic texture map which allows us to render video realistic controllable characters.
%
%
\paragraph{Differentiable Rendering and Neural Rendering}
Differentiable rendering bridges the gap between 2D supervision and unknown 3D scene parameters which one wants to learn or optimize.
Thus, differentiable rendering allows to train deep architectures, that learn 3D parameters of a scene, solely using 2D images for supervision. 
OpenDR~\cite{loper14a} first introduces an approximate differentiable renderer by representing a pixel as a linear combination of neighboring pixels and calculating pixel derivatives using differential filters. 
\cite{kato18} propose a 3D mesh renderer that is differentiable up to the visibility assumed to be constant during one gradient step. 
\cite{liu19a} differentiate through the visibility function and replace the z-buffer-based triangle selection with a probabilistic approach which assigns each pixel to all faces of a mesh. 
DIB-R~\cite{chen2019dibrender} propose to compute gradients analytically for all pixels in an image by representing foreground rasterization as a weighted interpolation of a face's vertex attributes and representing background rasterization as a distance-based aggregation of global face information. 
SDFDiff~\cite{jiang2020sdfdiff} introduces a differentiable renderer based on ray-casting SDFs. 
Our implementation of differentiable rendering follows the one of \cite{kato18} where we treat the surface as non transparent and thus the visibility is non-differentiable.
This is preferable in our setting as treating the human body and clothing as transparent would lead to wrong surface deformations and blurry dynamic textures.
%
\par
Different from differentiable rendering, neural rendering makes almost no assumptions about the physical model and uses neural networks to learn the rendering process from data to synthesize photo-realistic images. 
Some neural rendering methods ~\cite{chan2019dance, SiaroSLS2017, Pumarola_2018_CVPR, shysheya19, Martin18, Thies2019, kim18, liu19, liu20, Sarkar2020, yoon2020poseguided, MaSJSTV2017, Lischinski2018,Ma18} employ image-to-image translation networks~\cite{pix2pix2017, WangLZTKC2018, wang2018vid2vid} to augment the quality of the rendering.  
However, most of these methods suffer from view and/or temporal inconsistency. 
To enforce view and temporal consistency, some attempts were made to learn scene representations for novel view synthesis from 2D images.
Although this kind of methods achieve impressive renderings on static scenes~\cite{sitzmann2019deepvoxels,sitzmann2019srns,mildenhall2020nerf,liu2020neural,Zhang20arxiv_nerf++} and dynamic scenes for playback or implicit interpolation~\cite{lombardi2019neural,Zhang20arxiv_nerf++,Park20arxiv_nerfies,Pumarola20arxiv_D_NeRF,Tretschk20arxiv_NR-NeRF,Li20arxiv_nsff,Xian20arxiv_stnif,peng2020neural,raj2020pva,wang2020learning} and face \cite{gafni2020dynamic}, it is not straightforward to extend these methods to synthesize full body human images with explicit pose control. 
Instead, our approach can achieve video-realistic renderings of the full human body with motion- and view-dependent dynamic textures for arbitrary body poses \textit{and} camera views.
%
%
%
\begin{figure*}[t]
	\centering
	\includegraphics[width=0.90\textwidth]{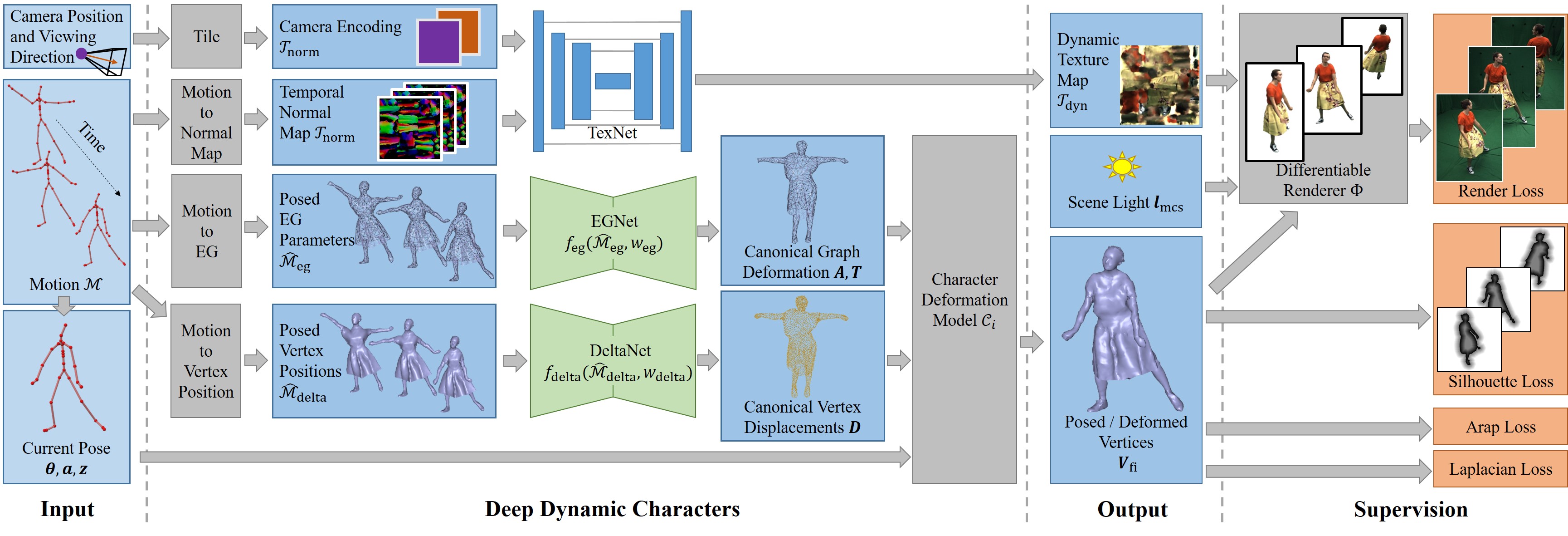} 
	\caption
	{
		Overview.
		Our method takes a motion sequence as input.
		We convert the pose information into task specific representations making the regression task easier for our network as input and output share the same representation.
		Then our two networks regress the motion-dependent coarse and fine deformations in the canonical pose.
		Given pose and deformations, our deformation layer outputs the posed and deformed character.
		Further, our TexNet regresses a motion- and view-dependent dynamic texture map.
		The regressed geometry as well as texture are weakly supervised based on multi-view 2D images.
	}
	\label{fig:overview}
\end{figure*}
%
%
\section{Method}
\label{sec:method}
%
%
%
Given multi-view images for training, our goal is to learn a poseable 3D character with dense deforming geometry of the full body and view- and motion-dependent textures that can be driven just by posing a skeleton and defining a camera view.
We propose a weakly supervised learning method with only multi-view 2D supervision in order to remove the need of detailed 3D ground truth geometry and 3D annotations.
Once trained, our network takes the current pose and a frame window of past motions of a moving person as input and outputs the motion-dependent geometry and texture, as shown in Fig.~\ref{fig:overview}. 
Note that our deformed geometry captures not only per-bone rigid transformations via classical skinning but also non-rigid deformations of clothing dependent on the current pose as well as the velocity and acceleration derived from the past motions.
In the following, we first introduce our deformable character model (Sec.~\ref{sec:characterModel}) and the data acquisition process (Sec.~\ref{sec:dataCapture}).
To regress the non-rigid deformations, we proceed in a coarse-to-fine manner.
First, we regress the deformation as rotations and translations of a coarse embedded graph (Sec.~\ref{sec:egnet}) only using multi-view foreground images as supervision signal.
As a result, we obtain a posed and deformed character that already matches the silhouettes of the multi-view images.
Next, we define a differentiable rendering layer which allows us to optimize the scene lighting which accounts for white balance shift and directional light changes (Sec~\ref{sec:lighting}).
Finally, our second network regresses per-vertex displacements to account for finer wrinkles and deformations that cannot be captured by the embedded deformation.
This layer can be trained using again the foreground masks, but in addition we also supervise it with a dense rendering loss using the previously optimized scene lighting (Sec.~\ref{sec:deltanet}).
Last, our dynamic texture network takes a view and motion encoding in texture space and outputs a dynamic texture (Sec.~\ref{sec:texture}) to further enhance the realism of our 3D character.
Similar to before, the texture network is weakly supervised using our differentiable renderer.
Note that none of our components requires ground truth 3D geometry and can be entirely trained weakly supervised.

%
%
\subsection{Character Deformation Model}
\label{sec:characterModel}
%
%
\paragraph{Acquisition}
Our method is person-specific and requires a 3D template model of the actor. 
We first scan the actor in T-pose using a 3D scanner~\cite{treedys}. 
Next, we use a commercial multi-view stereo reconstruction software~\cite{agisoftphotoscan} to reconstruct the 3D mesh with a static texture $\mathcal{T}_\mathrm{st}$ and downsample the reconstructed mesh to a resolution of around 5000 vertices. 
Like \cite{habermann19,habermann20}, we manually segment the mesh using the common human parsing labels and define per-vertex rigidity weights $s_i$ to model different degrees of deformation for different materials, where low rigidity weights allow more non-rigid deformations and vice versa, e.g., skin has higher rigidity weights than clothing.
%
%
\paragraph{Skeleton}
The template mesh is manually rigged to a skeleton. 
Here, the skeleton is parameterized as the set ${\mathcal{S}}= \{{\boldsymbol{\theta}},{\boldsymbol{\alpha}},{\mathbf{z}}\}$ with joint angles $\boldsymbol{\theta} \in \mathbb{R}^{57}$, global rotation $\boldsymbol{\alpha} \in \mathbb{R}^3$, and global translation $\mathbf{z} \in \mathbb{R}^3$, where skinning weights are automatically computed using Blender~\cite{blender}. 
This allows us to deform the mesh for a given pose by using dual quaternion skinning~\cite{kavan07}. 
%
%
\paragraph{Embedded Deformation}
As discussed before, the traditional skinning process alone is hardly able to model non-rigid deformations such as the swinging of a skirt.
To address this issue, we model the non-rigid deformations in the canonical pose from coarse to fine \emph{before} applying dual quaternion skinning.
On the coarse level, large deformations are captured with the embedded deformation representation~\cite{sorkine07,sumner07} which requires a small amount of parameters. 
We construct an embedded graph $\mathcal{G}$ consisting of $K$ nodes ($K$ is around 500 in our experiments) by downsampling the mesh.
The embedded graph $\mathcal{G}$ is parameterized with $\mathbf{A} \in \mathbb{R}^{K \times 3}$ and $\mathbf{T} \in \mathbb{R}^{K \times 3}$, where each row $k$ of $\mathbf{A}$ and $\mathbf{T}$ is the local rotation $\mathbf{a}_k\in \mathbb{R}^3$ in the form of Euler angles and local translation $\mathbf{t}_k\in \mathbb{R}^3$ of node $k$ with respect to the initial position $\mathbf{g}_k$ of node $k$. 
The connectivity of the graph node $k$ can be derived from the connectivity of the downsampled mesh and is denoted as $\mathcal{N}_\mathrm{n}(k)$.
To deform the original mesh with the embedded graph, the movement of each vertex on the original mesh is calculated as a linear combination of the movements of all the nodes of the embedded graph. 
Here, the weights $w_{i,k} \in \mathbb{R}$ for vertex $i$ and node $k$ are computed based on the geodesic distance between the vertex $i$ and the vertex on the original mesh that has the smallest Euclidean distance to the node $k$, where the weight is set to zero if the distance exceeds a certain threshold. 
We denote the set of nodes that finally influences the movement of the vertex $i$ as $\mathcal{N}_{\mathrm{vn}}(i)$.
%
%
\paragraph{Vertex Displacements}
On the fine level, in addition to the embedded graph, which models large deformations, we use vertex displacements to recover fine-scale deformations, where a displacement $\mathbf{d}_i \in \mathbb{R}^3$ is assigned to each vertex $i$. 
Although regressing so many parameters is not an easy task, the training of the vertex displacement can still be achieved since the embedded graph captures most of deformations on a coarse level.
Thus, the regressed displacements, the network has to learn, are rather small.
%
%
\paragraph{Character Deformation Model}
Given the skeletal pose $\boldsymbol{\theta}$, $\boldsymbol{\alpha}$, $\mathbf{z}$, the embedded graph parameters $\mathbf{A}$, $\mathbf{T}$, and the vertex displacements $\mathbf{d}_i$, we can deform each vertex $i$ with the function
%
%
\begin{equation} \label{eq:character}
 \mathcal{C}_i(\boldsymbol{\theta}, \boldsymbol{\alpha}, \mathbf{z}, \mathbf{A}, \mathbf{T}, \mathbf{d}_i) = \mathbf{v}_i
\end{equation}
%
%
which defines our final character representation.
Specifically, we first apply the embedded deformation and the per-vertex displacements to the template mesh in canonical pose, which significantly simplifies the learning of non-rigid deformations by alleviating ambiguities in the movements of mesh vertices caused by pose variations.
Thus, the deformed vertex position is given as
%
%
\begin{equation} \label{eq:vnr}
\mathbf{y}_{i}= 
\mathbf{d}_i + \sum_{k \in \mathcal{N}_{\mathrm{vn}}(i)}
w_{i,k}
(
R(\mathbf{a}_k)(\hat{\mathbf{v}}_i-\mathbf{g}_k) +\mathbf{g}_k + \mathbf{t}_k
),
\end{equation}
%
%
where $\hat{\mathbf{v}}_i$ is the initial position of vertex $i$ in the template mesh.
$R:\mathbb{R}^3 \rightarrow SO(3)$ converts the Euler angles to a rotation matrix.
We apply the skeletal pose to the deformed vertex $\mathbf{y}_{i}$ in canonical pose to obtain the deformed and posed vertex in the global space
%
%
\begin{equation} \label{eq:vi}
\mathbf{v}_{i} = \mathbf{z} +
\sum_{k \in \mathcal{N}_{\mathrm{vn}}(i)}
w_{i,k}
(
R_{\mathrm{sk},k}(\boldsymbol{\theta}, \boldsymbol{\alpha}) \mathbf{y}_{i} + t_{\mathrm{sk},k}(\boldsymbol{\theta}, \boldsymbol{\alpha})
),
\end{equation}
%
where the rotation $R_{\mathrm{sk},k}$ and the translation $t_{\mathrm{sk},k}$ are derived from the skeletal pose using dual quaternion skinning, and $\mathbf{z}$ is the global translation of the skeleton. 
Note that Eq.~\ref{eq:vnr} and \ref{eq:vi} are fully differentiable with respect to pose, embedded graph and vertex displacements.
Thus, gradients can be propagated in learning frameworks.
The final model does not only allow us to pose the mesh via skinning but also to model non-rigid surface deformations in a coarse to fine manner via embedded deformation and vertex displacements.
Further, it disentangles the pose and the surface deformation, where the later is represented in the canonical pose space. 
%
%
\par 
The main difference to data-driven body models, e.g. SMPL~\cite{loper15}, is that our character formulation allows posing, deforming, and texturing using an effective and simple equation which is differentiable to all its input parameters. 
SMPL and other human body models do not account for deformations (e.g. clothing) and they also do not provide a texture. 
Our specific formulation allows seamless integration into a learning framework and learning its parameters conditioned on skeletal motion (and camera pose) as well as adding spatial regularization from coarse to fine which is important when considering a weakly supervised setup.
%
%
\subsection{Data Capture and Motion Preprocessing}
\label{sec:dataCapture}
For supervision, our method requires multi-view images and foreground masks of the actor performing a wide range of motions at varying speeds to sample different kinds of dynamic deformations of clothing caused by the body motion.
Thus, we place the subject in a multi camera capture studio with green screen and record a sequence with $C=120$ synchronized and calibrated 4K cameras at 25 frames per second.
We apply color keying to segment the foreground and convert the foreground masks to distance transform images~\cite{borgefors86}.
We denote the $f$th frame of camera $c$ and its corresponding distance transform image and the foreground mask as $\mathcal{I}_{c,f}$, $\mathcal{D}_{c,f}$, and $\mathcal{F}_{c,f}$, respectively.
%
%
\begin{figure*}[t]
	\centering
	\includegraphics[width=0.95\textwidth]{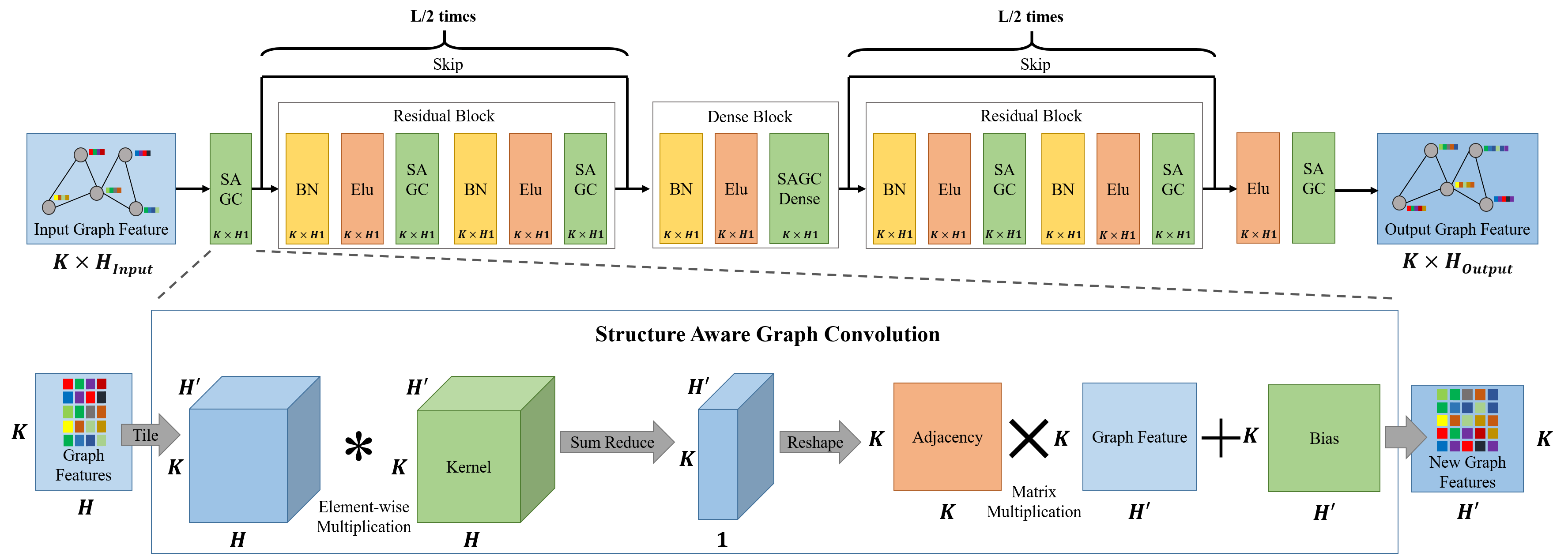} 
	\caption
	{
		Structure aware graph convolutional network (top) as well as a detailed illustration of the proposed Structure Aware Graph Convolution (bottom).
	}
	\label{fig:gcnnOverview}
	\vspace{-12pt}
\end{figure*}
%
%
\par 
We further track the human motions using a multi-view markerless motion capture system~\cite{captury}. 
We denote the tracked motion of the $f$th frame as ${\mathcal{S}}_{f}= \{{\boldsymbol{\theta}}_f,{\boldsymbol{\alpha}}_f,{\mathbf{z}}_f\}$. 
Next, we normalize the temporal window of motions ${\mathcal{M}_t}=\{{\mathcal{S}}_{f}: f \in \{t-F,..., t\}\}$ for geometry and texture generation separately as it is very hard to sample all combinations of rigid transforms and joint angle configurations during training data acquisition.
The normalization for geometry generation is based on two observations: 
1) The global position of the motion sequence should not influence the dynamics of the geometry; we therefore normalize the global translation of ${\mathcal{S}}_{f}$ across different temporal windows of motions while keeping relative translations between the frames in each temporal window, i.e., we set $\hat{\mathbf{z}}_{t}=\mathbf{0}$ and $\hat{\mathbf{z}}_{t'}={\mathbf{z}}_{t'} - {\mathbf{z}}_{t}$ for $t' \in \{t-F, ...,t-1\}$ where $\mathbf{0}$ is the zero vector in $\mathbb{R}^3$.
2) The rotation around the $y$ axis will not affect the geometry generation as it is in parallel with the gravity direction; thus, similar to normalizing the global translation, we set $\hat{\boldsymbol{\alpha}}_{y, t}=0$ and $\hat{\boldsymbol{\alpha}}_{y, t'}={\boldsymbol{\alpha}}_{y, t'} - {\boldsymbol{\alpha}}_{y, t}$. 
We denote the temporal window of the normalized motions for geometry generation as $\hat{\mathcal{M}_t}=\{\hat{\mathcal{S}}_{f}: f \in \{t-F,..., t\}\}$, where $\hat{\mathcal{S}}_{f}= \{\hat{\boldsymbol{\theta}}_f,\hat{\boldsymbol{\alpha}}_f,\hat{\mathbf{z}}_f\}$.
For texture generation, we only normalize the global translation, but not the rotation around the $y$ axis to get the normalized motions $\tilde{\mathcal{M}}_{t}$, since we would like to generate view-dependent textures where the subjects relative direction towards the light source and therefore the rotation around the $y$ axis matters.
In all our results, we set $F=2$.
For readability reasons, we assume $t$ is fixed and drop the subscript.

%
%
\subsection{Embedded Deformation Regression}
\label{sec:egnet}
%
%
\subsubsection{Embedded Deformation Regression}
\label{sec:embeddeddeformationregression}
With the skeletal pose alone, the non-rigid deformations of the skin and clothes cannot be generated. 
We therefore introduce an embedded graph network, \textit{EGNet}, to produce coarse-level deformations.
\textit{EGNet} learns a mapping from the temporal window of normalized motions $\hat{\mathcal{M}}$ to the rotations $\mathbf{A}$ and translations $\mathbf{T}$ of the embedded graph defined in the canonical pose for the current frame (i.e., the last frame of the temporal window).
\textit{EGNet} learns deformations correlated to the velocity and acceleration at the current frame since it takes the pose of the current frame as well as the previous two frames as input.
Directly regressing $\mathbf{A}$ and $\mathbf{T}$ from the normalized skeletal motion $\hat{\mathcal{M}}$ is challenging as the input and output are parameterized in a different way, i.e., $\hat{\mathcal{M}}$ represents skeleton joint angles while $\mathbf{A}$ and $\mathbf{T}$ model rotation and translation of the graph nodes.
To address this issue, we formulate this regression task as a graph-to-graph translation problem rather than a skeleton-to-graph one. 
Specifically, we pose the embedded graph with the normalized skeletal motion $\hat{\mathcal{M}}$ using dual quaternion skinning~\cite{kavan07} to obtain the rotation and translation parameters $\hat{\mathcal{M}}_\mathrm{eg} \in \mathbb{R}^{K \times 6(F+1)}$ of the embedded graph. 
Therefore, the mapping of \textit{EGNet} can be formulated as $f_\mathrm{eg}(\hat{\mathcal{M}}_\mathrm{eg},\mathbf{w}_\mathrm{eg})=(\mathbf{A},\mathbf{T})$, which takes the posed embedded graph rotations and translations $\hat{\mathcal{M}}_\mathrm{eg}$ and learnable weights $\mathbf{w}_\mathrm{eg}$ as inputs and outputs the embedded deformation $(\mathbf{A},\mathbf{T})$ in canonical pose. 
Using the character representation defined in Eq.~\ref{eq:character}, the posed and coarsely deformed character is defined as
%
%
\begin{equation}\label{eq:coarseCharacter}
	\mathcal{C}_i(\boldsymbol{\theta}, \boldsymbol{\alpha}, \mathbf{z}, f_\mathrm{eg}(\hat{\mathcal{M}}_\mathrm{eg},\mathbf{w}_\mathrm{eg}), \mathbf{0}) = \mathbf{v}_{\mathrm{co},i}.
\end{equation}
%
%
Here, $\boldsymbol{\theta}, \boldsymbol{\alpha}, \mathbf{z}$ are the unnormalized pose of the last frame of the motion sequence and the displacements are set to zero.
Next, we explain our novel graph convolutional architecture of EGNet.
%
%
\subsubsection{Structure Aware Graph Convolution}
\label{sec:graphConvOperator}
Importantly, our graph is fixed as our method is person specific.
Thus, the spatial relationship between the graph nodes and their position implicitly encode a strong prior.
For example, a node that is mostly attached to skin vertices will deform very different than nodes that are mainly connected to vertices of a skirt region.
This implies that learnable node features require different properties depending on which node is considered.
However, recent graph convolutional operators \cite{defferrard17} apply the same filter on every node which contradicts the above requirements.
Therefore, we aim for a graph convolutional operator that applies an individual kernel per node.
%
%
\par
Thus, we propose a new \textit{Structure Aware Graph Convolution (SAGC)}.
To define the per node SAGC, we assume an input node feature $\mathbf{f}_k \in \mathbb{R}^H$ of size $H$ is given and the output feature dimension is $H'$ for a node $k$.
Now, the output feature $\mathbf{f}'_k$ can be computed as
%
%
\begin{equation}
	\mathbf{f}'_k
	=
	\mathbf{b}_k + \sum_{l \in \mathcal{N}_R(k)}
	a_{k,l} \mathbf{K}_l \mathbf{f}_l
\end{equation}
%
%
where $\mathcal{N}_R(k)$ is the $R$-ring neighbourhood of the graph node $k$.
$\mathbf{b}_k \in \mathbb{R}^{H'}$ and $\mathbf{K}_l \in \mathbb{R}^{H' \times H}$ are a trainable bias vector and kernel matrix.
$a_{k,l}$ is a scalar weight that is computed as 
%
%
\begin{equation}
a_{k,l} =
\frac
{r_{k,l}}
{ \sum_{l \in \mathcal{N}_R(k)} r_{k,l}}
\end{equation}
%
%
where for a node $k$ $r_{k,l}$ is the inverse ring value, e.g., for the case $l=k$ the value is $R$ and for the direct neighbours of $k$ the value is $R-1$.
More intuitively, our operator computes a linear combination of modified features $\mathbf{K}_l \mathbf{f}_l$ of node $k$ and neigbouring nodes $l$ within the $R$-ring neighbourhood weighted by $a_{k,l}$ that has a linear falloff to obtain the new feature for node $k$.
Importantly, each node has its own learnable kernel $\mathbf{K}_l$ and bias $\mathbf{b}_k$ weights allowing features at different locations in the graph to account for different spatial properties.
As shown at the bottom of Fig.~\ref{fig:gcnnOverview}, the features for each node can be efficiently computed in parallel and by combining all the per node input/output features, one obtains the corresponding input/output feature matrices $\mathbf{F}_k, \mathbf{F}'_k$.
%
%
\subsubsection{Structure Aware Graph Convolutional Network}
\label{sec:egnetwork}
Our structure-aware graph convolutional network (SAGCN) takes as input a graph feature matrix $\mathbf{F}_\mathrm{input} \in \mathbb{R}^{K \times H_\mathrm{input}}$ and outputs a new graph feature matrix $\mathbf{F}_\mathrm{output} \in \mathbb{R}^{K \times H_\mathrm{output}}$ (see Fig.~\ref{fig:gcnnOverview}).
First, $\mathbf{F}_\mathrm{input}$ is convolved with the SAGC operator, resulting in a feature matrix of size $K \times H_\mathrm{1}$.
Inspired by the ResNet architecture~\cite{he16}, we also use so-called residual blocks that take the feature matrix of size $K \times H_\mathrm{1}$ and output a feature matrix of the same size.
Input and output feature matrices are connected via skip connections which prevent vanishing gradients, even for very deep architectures.
A residual block consists of two chains of a batch normalization, an Elu activation function, and a SAGC operation.
For a very large number of graph nodes, the local features can barely spread through the entire graph.
To still allow the network to share features between far nodes we propose a so-called dense block consisting of a batch normalization, an Elu activation, and a SAGC operator. 
Importantly, for this specific dense block we set all $a_{k,l}=1$ which allows to share features between far nodes.
In total, we use $L$ residual blocks, half of them before and half of them after the dense block. 
The last layers (Elu and SAGC) resize the features to the desired output size.
\par 
We now define EGNet as a SAGCN architecture where the graph is given as the embedded graph $\mathcal{G}$.
The input feature matrix is given by the normalized embedded graph rotations and translations $\hat{\mathcal{M}}_\mathrm{eg}$ and the output is the deformation parameters $(\mathbf{A},\mathbf{T})$.
Thus, the input and output feature sizes are $H_\mathrm{input}=6(F+1)$ and $H_\mathrm{output}=6$, respectively.
Further, we set $H_1=16$, $L=8$, and $R=3$.
As $\mathcal{G}$ only contains around 500 nodes, we did not employ a dense block.
%
%
\subsubsection{Weakly Supervised Losses}
\label{sec:egnetlosses}
To train the weights $\mathbf{w}_\mathrm{eg}$ of EGNet $f_\mathrm{eg}(\hat{\mathcal{M}}_\mathrm{eg},\mathbf{w}_\mathrm{eg})$, we only employ a weakly supervised loss on the posed and deformed vertices $\mathbf{V}_{\mathrm{co}}$ and on the regressed embedded deformation parameters $(\mathbf{A},\mathbf{T})$ directly as
%
%
\begin{equation} 
\mathcal{L}_{\mathrm{eg}}(\mathbf{V}_{\mathrm{co}},\mathbf{A},\mathbf{T}) =\mathcal{L}_{\mathrm{sil}}(\mathbf{V}_{\mathrm{co}}) + \mathcal{L}_{\mathrm{arap}}(\mathbf{A},\mathbf{T}).
\end{equation}
%
%
Here, the first term is our multi-view image-based data loss and the second term is a spatial regularizer.
%
%
\paragraph{Silhouette Loss.}
Our multi-view silhouette loss
%
%
\begin{equation} \label{eq:lossSil}
\begin{split}
\mathcal{L}_{\mathrm{sil}}(	\mathbf{V}_{\mathrm{co}}) &= 
\sum_{c=1}^C 
\sum_{i \in \mathcal{B}_c} 
\rho_{c,i}
\|\mathcal{D}_{c}\left(
\pi_c\left(	
\mathbf{v}_{\mathrm{co},i}
\right)
\right)
\|^2 \\
&+
\sum_{c=1}^C 
\sum_{\mathbf{p} \in \{\mathbf{u} \in \mathbb{R}^2| \mathcal{D}_c(\mathbf{u})=0\}}
\|
\pi_c\left(	
\mathbf{v}_{\mathrm{co},\mathbf{p}}
\right)
-
\mathbf{p}
\|^2
\end{split}
\end{equation}
%
%
ensures that the silhouette of the projected character model aligns with the multi-view image silhouettes in an analysis-by-synthesis manner.
Therefore, we employ a bi-sided loss.
The first part of Eq.~\ref{eq:lossSil} is a model-to-data loss which enforces that the projected boundary vertices are pushed to the zero contour line of the distance transform image $\mathcal{D}_c$ for all cameras $c$.
Here, $\pi_c$ is the perspective camera projection of camera $c$ and $\rho_{c,i}$ is a scalar weight accounting for matching image and model normals~\cite{habermann19}.
$\mathcal{B}_c$ is the set of boundary vertices, e.g., the vertices that lie on the boundary after projecting onto camera view $c$.
$\mathcal{B}_c$ can be efficiently computed using our differentiable renderer, that is introduced later, by rendering out the depth maps and checking if a projected vertex lies near a background pixel in the depth map. 
The second part of Eq.~\ref{eq:lossSil} is a data-to-model loss which ensures that all silhouette pixels $\{\mathbf{u} \in \mathbb{R}^2| \mathcal{D}_c(\mathbf{u})=0\}$ are covered by their closest vertex $\mathbf{v}_{\mathrm{co},\mathbf{p}}$ using the Euclidean distance in 2D image space as the distance metric.
%
%
\paragraph{ARAP Loss.}
Only using the above loss would lead to an ill-posed problem as vertices could drift along the visual hull carved by the silhouette images without receiving any penalty resulting in distorted meshes.
Thus, we employ an as-rigid-as-possible regularization~\cite{sorkine07,sumner07} defined as 
%
%
\begin{equation} \label{eq:lossARAP}
\mathcal{L}_\mathrm{arap}(\mathbf{A},\mathbf{T}) = 
\sum_{k=1}^K 
\sum_{l \in \mathcal{N}_\mathrm{n}(k)}
u_{k,l}
\lVert 
d_{k,l}(\mathbf{A},\mathbf{T}) 
\rVert_1
\end{equation}
%
%
$$
d_{k,l}(\mathbf{A},\mathbf{T})\! \!= \!\!
R(\mathbf{a}_k) (\mathbf{g}_l - \mathbf{g}_k) + \mathbf{t}_k + \mathbf{g}_k
- (\mathbf{g}_l + \mathbf{t}_l).
$$
%
%
We use material aware weighting factors $u_{k,l}$~\cite{habermann20} computed by averaging the rigidity weights $s_i$ of all vertices attached to node $k$ and $l$.
Thus, different levels of rigidity are assigned to individual surface parts, e.g., graph nodes attached to skirt vertices can deform more freely than those attached to skin vertices.
%
%
\subsection{Lighting Estimation}
\label{sec:lighting}
\begin{figure}[t]
	\centering
	\includegraphics[width=\linewidth]{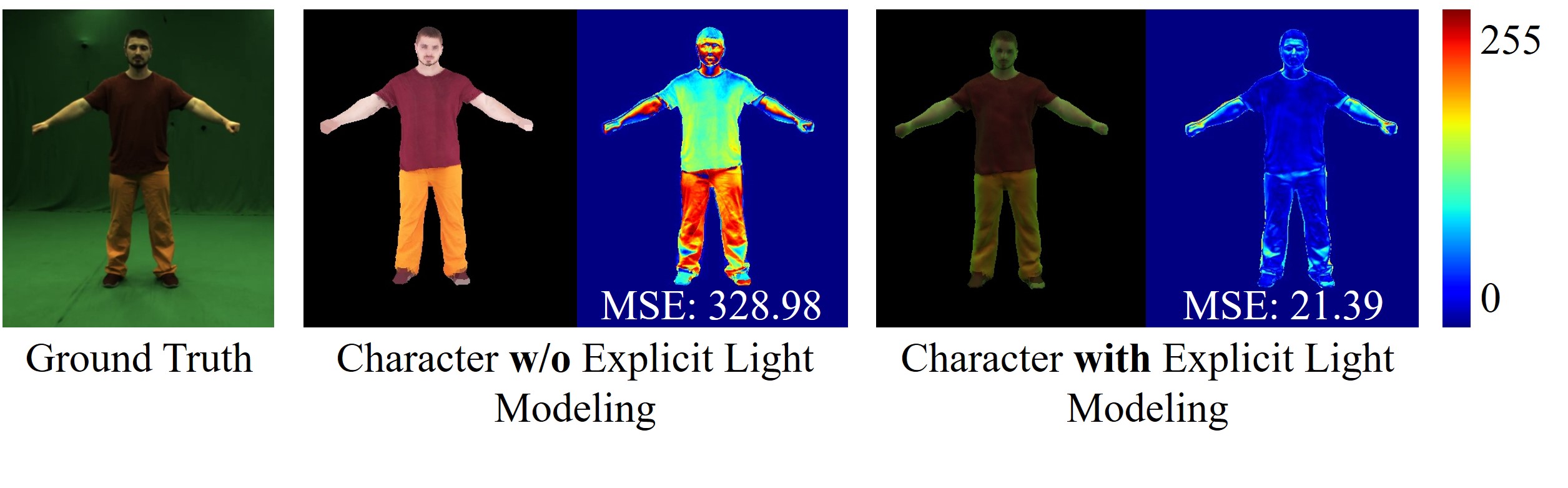} 
	\caption
	{
	    From left to right. 
	    Ground truth image. 
	    Template rendered with the static texture. 
	    Mean squared pixel error (MSE) image computed from the masked ground truth and the rendering.
	    Template rendered with the static texture and \textit{the optimized lighting}.
	    MSE image computed from the ground truth and the render with optimized lighting.
	    Note that the optimized lighting clearly lowers the error between the rendering and the ground truth which is advantageous for learning the per-vertex displacements.
	    Further note that this is not our final appearance result.
	    Instead, we only use the optimized lighting to improve the dense rendering loss which supervises our displacement network that will be introduced in the next section.
	}
	\label{fig:lighting}
\end{figure}

So far, we can obtain the posed and coarsely deformed character $\mathbf{V}_{\mathrm{co}}$ using EGNet and Eq.~\ref{eq:coarseCharacter}.
What is still missing are the finer deformations which are hard to capture just with multi-view silhouette images.
Thus, we aim for a dense rendering loss that takes the posed and deformed geometry along with the static texture $\mathcal{T}_\mathrm{st}$, renders it into all camera views and compares it to the corresponding images.
However, the lighting condition differs when capturing the scan of the subject and therefore the texture and the lighting in the multi-camera studio can vary due to different light temperatures, camera optics and sensors, and scene reflections as shown in Fig.~\ref{fig:lighting}.
As a remedy, we propose a differentiable rendering that also accounts for the difference in lighting and explicitly optimize the lighting parameters for the multi-camera studio sequences. 
%
%
%
%
%
\paragraph{Differentiable Rendering.}
We assume the subject has a purely Lambertian surface reflectance~\cite{lambert1760} and that the light sources are sufficiently far away resulting in an overall smooth lighting environment.
Hence, we can use the efficient Spherical Harmonics (SH) lighting representation~\cite{mueller66} which models the scene lighting only with a few coefficients.
To account for view-dependent effects, each of the $C$ cameras has its own lighting coefficients $\mathbf{l}_\mathrm{c} \in \mathbb{R}^{9 \times 3}$ which in total sums up to $27C$ coefficients.
We assume that the image has a resolution of $W \times H$.
To compute the RGB color of a pixel $\mathbf{u} \in \mathcal{R}$ where $\mathcal{R} = \{ (u,v) | u \in [1,W],v \in [1,H] \}$ in camera view $c$, we use the rendering function 
%
%
\begin{equation}\label{eq:renderingfunction}
	\Phi_{c,\mathbf{u}}(\mathbf{V}, \mathcal{T}, \mathbf{l}_\mathrm{c}) = a_{c,\mathbf{u}}(\mathbf{V},\mathcal{T}) 
	\cdot
	i_{c,\mathbf{u}}(\mathbf{V},\mathbf{l}_\mathrm{c})
\end{equation}
%
%
which takes the vertex positions $\mathbf{V}$, the texture $\mathcal{T}$, and the lighting coefficients $\mathbf{l}_\mathrm{c}$ for camera $c$.
As we assume a Lambertian reflectance model, the rendering equation simplifies to a dot product of the albedo $a_{c,\mathbf{u}}(\mathbf{V},\mathcal{T})$ of the projected surface and the illumination $i_{c,\mathbf{u}}(\mathbf{V},\mathbf{l}_\mathrm{c})$.
The albedo can be computed as 
%
%
\begin{equation}
	a_{c,\mathbf{u}}(\mathbf{V},\mathcal{T}) =
	v_{c,\mathbf{u}}(\mathbf{V}) t_{c,\mathbf{u}}(\mathbf{V},\mathcal{T}) 
\end{equation}
%
%
where $v_{c,\mathbf{u}}(\mathbf{V})$ is an indicator function that computes whether a surface is visible or not given the pixel position, camera, and surface.
Like traditional rasterization~\cite{pineda88}, $t_{c,\mathbf{u}}(\mathbf{V},\mathcal{T})$ computes the barycentric coordinates of the point on the triangle that is covering pixel $\mathbf{u}$, which are then used to bi-linearly sample the position in texture map space.
The lighting can be computed in SH space as 
%
%
\begin{equation}
i_{c,\mathbf{u}}(\mathbf{V},\mathbf{l}_\mathrm{c}) =
\sum_{j=1}^9
\mathbf{l}_{c,j} SH_j(n_{c,\mathbf{u}}(\mathbf{V}))
\end{equation}
%
%
where $\mathbf{l}_{c,j} \in \mathbb{R}^3$ are the $j$th SH coefficients for each color channel and $SH_j$ are the corresponding SH basis functions.
$n_{c,\mathbf{u}}(\mathbf{V})$ computes the screen space pixel normal given the underlying geometry.
\par 
Note that the final color $\Phi_{c,\mathbf{u}}(\mathbf{V}, \mathcal{T}, \mathbf{l}_\mathrm{c})$ only depends on the geometry $\mathbf{V}$, the texture $\mathcal{T}$, and the lighting coefficients $\mathbf{l}_\mathrm{c}$ assuming camera and pixel position are fixed.
As all above equations (except visibility) are differentiable with respect to these variables we can backpropagate gradients through the rendering process.
The visibility $v_{c,\mathbf{u}}(\mathbf{V})$ is fixed during one gradient step.
%
%
\paragraph{Lighting Optimization.}
To optimize the lighting, we assume the texture and geometry are fixed.
Therefore, we set the texture to $\mathcal{T}=\mathcal{T}_\mathrm{st}$ which is the static texture obtained from the scan.
For the geometry, we set $\mathbf{V}=\mathbf{V}_\mathrm{co}$ which is the deformed and posed vertex positions regressed by EGNet.
Now, the lighting coefficients $\mathbf{l}_{\mathrm{mcs},c}$ for camera $c$ can be computed by minimizing
%
%
\begin{equation}\label{eq:losslight}
\mathcal{L}_\mathrm{light}(\mathbf{l}_{\mathrm{mcs},c})
=
\sum_{\mathbf{u} \in \mathcal{R}}
\|
	\Phi_{c,\mathbf{u}}(\mathbf{V}_\mathrm{co}, \mathcal{T}_\mathrm{st}, \mathbf{l}_{\mathrm{mcs},c})
	-
	\mathcal{I}_{c,\mathbf{u}}
\|^2
\end{equation}
%
%
for all frames of the training sequence.
Note that we use all frames while the lighting coefficients are the same across frames.
As we cannot solve for all frames jointly, we employ a stochastic gradient descent which randomly samples 4 frames and apply $30{,}000$ iterations.
As a result, the optimal lighting coefficients $\mathbf{l}^*_{\mathrm{mcs},c}$ are obtained and the rendering with the static texture and the optimized lighting matches the global appearance much better than a rendering which is not explicitly modeling lighting (see Fig.~\ref{fig:lighting}).
As the lighting and the texture are now known, we can leverage the rendering function Eq.~\ref{eq:renderingfunction} to densely supervise the per-vertex displacements which we want to regress on top of the embedded deformation parameters. 
%
%
\subsection{Vertex Displacement Regression}
\label{sec:deltanet}
%
%
\subsubsection{Displacement Network DeltaNet}
\label{sec:deltanetwork}
\begin{figure}[t]
	\centering
	\includegraphics[width=\linewidth]{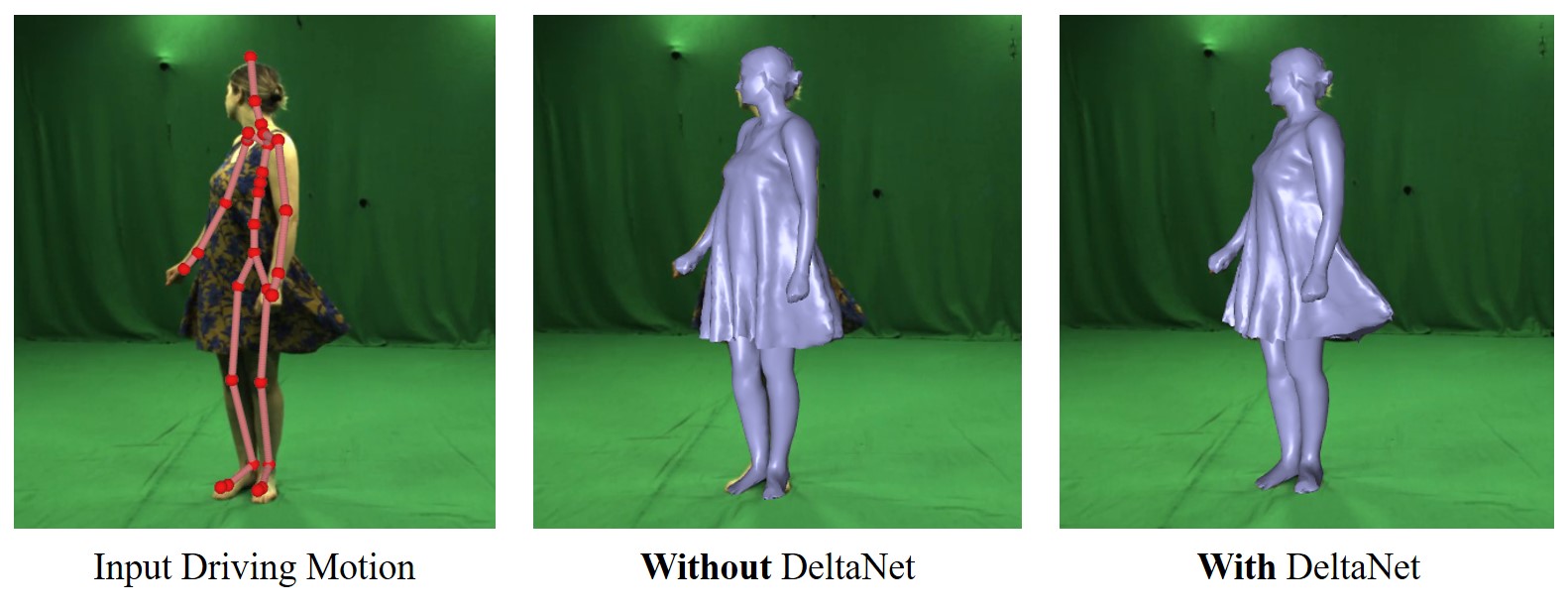} 
	\caption
	{
		From left to right. 
		Input motion. 
		Our result without the displacements predicted by DeltaNet. 
		Our result with the predicted displacements.
		Note that the displacements clearly improve the overlay as they allow capturing finer geometric details.
	}
	\label{fig:withWithoutDelta}
\end{figure}
Our goal is capturing also finer deformations which our character representation models as per-vertex displacements $\mathbf{d}_i$ that were previously set to zero.
Our second network, called \textit{DeltaNet}, takes again the motion sequence and regresses the displacements $\mathbf{D} \in \mathbb{R}^{N \times 3}$ for the $N$ vertices of the template mesh in canonical pose. 
Here, the $i$th row of $\mathbf{D}$ contains the displacement $\mathbf{d}_i$ for vertex $i$.
Similar to the EGNet, we represent the pose in the same space as the output space of the regression task.
Thus, we pose the template mesh to the respective poses from our normalized motion $\hat{\mathcal{M}}$ using dual quaternion skinning resulting in $F+1$ consecutive 3D vertex positions which we denote as $\hat{\mathcal{M}}_\mathrm{delta} \in \mathbb{R}^{N \times 3(F+1)}$.
We denote DeltaNet as the function $f_\mathrm{delta}(\hat{\mathcal{M}}_\mathrm{delta}, \mathbf{w}_\mathrm{delta})=\mathbf{D}$ where $\mathbf{w}_\mathrm{delta}$ are the trainable network weights.
Similarly, the displacement for a single vertex is referred to as $f_{\mathrm{delta},i}(\hat{\mathcal{M}}_\mathrm{delta}, \mathbf{w}_\mathrm{delta})=\mathbf{d}_i$ and the final posed and deformed character vertices are defined as 
\begin{equation}\label{eq:fineCharacter}
\mathcal{C}_i(\boldsymbol{\theta}, \boldsymbol{\alpha}, \mathbf{z}, f_\mathrm{eg}(\hat{\mathcal{M}}_\mathrm{eg},\mathbf{w}_\mathrm{eg}), f_{\mathrm{delta},i}(\hat{\mathcal{M}}_\mathrm{delta}, \mathbf{w}_\mathrm{delta})) = \mathbf{v}_{\mathrm{fi},i}.
\end{equation}
$\mathbf{V}_{\mathrm{fi}} \in \mathbb{R}^{N \times 3}$ denotes the matrix that contains all the posed and deformed vertices.
Again, we use our SAGCN architecture as it is able to preserve local structures better than fully connected architectures.
The graph is defined by the connectivity of our template mesh and each vertex is a graph node.
The input is $\hat{\mathcal{M}}_\mathrm{delta}$ and therefore the input and output feature sizes are $H_\mathrm{input}=3(F+1)$ and $H_\mathrm{output}=3$, respectively.
Further, we set $H_1=16$, $L=8$, and $R=3$.
Different to EgNet, we employ the dense block as the mesh graph is very large and thus sharing features for very far nodes is difficult otherwise.
Fig.~\ref{fig:withWithoutDelta} shows that adding these displacements improves the silhouette matching as the finer geometric details can be captured. 
%
%
\subsubsection{Weakly Supervised Losses}
\label{sec:deltanetlosses}
We weakly supervise the displacement predictions and therefore $\mathbf{V}_{\mathrm{fi}}$ using the loss function
\begin{equation} 
\mathcal{L}_{\mathrm{Delta}}(\mathbf{V}_{\mathrm{fi}}) =\mathcal{L}_{\mathrm{chroma}}(\mathbf{V}_{\mathrm{fi}}) + \mathcal{L}_{\mathrm{sil}}(\mathbf{V}_{\mathrm{fi}}) + \mathcal{L}_{\mathrm{lap}}(\mathbf{V}_{\mathrm{fi}})
\end{equation}
which is composed of two multi-view image-based data terms and a spatial regularizer.
$\mathcal{L}_{\mathrm{sil}}$ is the silhouette loss introduced in Eq.~\ref{eq:lossSil} but now applied to the vertices after adding the displacements to still ensure matching model and image silhouettes.
%
%
\paragraph{Chroma Loss}
The silhouette-based loss alone can only constrain the boundary vertices of the model.
But since we want to learn the displacements per vertex a denser supervision is required and therefore we employ a dense rendering loss 
%
%
\begin{equation}
\mathcal{L}_\mathrm{chroma}(\mathbf{V}_{\mathrm{fi}}) 
=
\sum_{c=1}^{C}
\sum_{\mathbf{u} \in \mathcal{R}}
\|
g(\Phi_{c,\mathbf{u}}(\mathbf{V}_{\mathrm{fi}}, \mathcal{T}_\mathrm{st}, \mathbf{l}^*_{\mathrm{mcs},c}))
-
g(\mathcal{I}_{c,\mathbf{u}})
\|^2
\end{equation}
%
%
which renders the mesh $\mathbf{V}_{\mathrm{fi}}$ into the camera view $c$ and compares it with the ground truth image $\mathcal{I}_c$ by using our differentiable renderer proposed in the previous section.
In contrast to the previous rendering loss (see Eq.~\ref{eq:losslight}), we apply the color transform $g$ to both the rendered and the ground truth image.
$g$ converts RGB values into the YUV color space and only returns the UV channels.
Thus, our loss is more invariant to shadow effects such as self shadows which cannot be modeled by our renderer.
Instead, the loss mainly compares the chroma values of the rendering and the ground truth.
%
%
\paragraph{Laplacian Loss}
Only using the multi-view image-based constrains can still lead to distorted geometry.
Thus, we further regularize the posed and deformed model with a Laplacian regularizer 
%
%
\begin{equation} \label{eq:lossLaplace}
\mathcal{L}_{\mathrm{lap}}(\mathbf{V}_{\mathrm{fi}}) = 
\sum_{i=1}^N 
s_i
\lVert
|\mathcal{N}_i| \left(	 \mathbf{v}_{\mathrm{fi},i} - \mathbf{v}_{\mathrm{co},i} \right)
-
\sum_{j \in \mathcal{N}_i} 
\left(	
\mathbf{v}_{\mathrm{fi},j} -\mathbf{v}_{\mathrm{co},j}
\right)
\rVert^2
\end{equation}
%
%
which ensures that the Laplacian of the mesh before and after adding the displacements are locally similar.
Here, $\mathcal{N}_i$ is the set that contains the indices of the one ring neighbourhood of vertex $i$ and $s_i$ are the per-vertex spatially varying regularization weights.
%
%

%
%
\subsection{Dynamic Texture Regression}
\label{sec:texture}
%
%
To add further realism to our poseable neural character it is indispensable to have a realistic looking texture.
Although our scan provides a static texture, we found that wrinkles are baked in and thus look unrealistic for certain poses and further it cannot account for view-dependent effects.
Therefore, our goal is to also regress a motion and view point dependent texture $\mathcal{T}_\mathrm{dyn}\in \mathbb{R}^{1024 \times 1024 \times 3}$.
%
%
\par 
As explained in Sec.~\ref{sec:dataCapture}, we use our normalized motion $\tilde{\mathcal{M}}$ as a conditioning input.
Regressing textures just from these joint angles is difficult as the input and output are in different spaces.
Thus, we pose the mesh according to the poses in $\tilde{\mathcal{M}}$ and render the global normals into a texture map.
By stacking the normal maps for each of the $F+1$ poses in $\tilde{\mathcal{M}}$, we obtain $\mathcal{T}_\mathrm{norm} \in \mathbb{R}^{1024 \times 1024 \times 3(F+1)}$ where we use a texture size of $1024 \times 1024$.
As textural appearance does not only depend on poses but also on the positioning of the subject with respect to the camera, we further encode the camera position and orientation into texture space denoted as $\mathcal{T}_\mathrm{cam}\in \mathbb{R}^{1024 \times 1024 \times 6}$ where each pixel contains the position and orientation of the camera.
By concatenating $\mathcal{T}_\mathrm{norm}$ and $\mathcal{T}_\mathrm{cam}$, we obtain $\mathcal{T}_\mathrm{input} \in \mathbb{R}^{1024 \times 1024 \times 3(F+1) + 6}$ which is our final input to the texture regression network.
%
%
\par 
Our texture network, \textit{TexNet}, is based on the UNet architecture~\cite{isola18} which we adapted to handle input and output dimensions of size $1024 \times 1024$.
It takes the input texture encoding $\mathcal{T}_\mathrm{input}$ and outputs the dynamic texture $\mathcal{T}_\mathrm{dyn}$.
Note that due to our input representation, the network can learn motion-dependent texture effects as well as view-dependent effects.
%
%
\paragraph{Photometric Loss}
To supervise $\mathcal{T}_\mathrm{dyn}$ for camera $c'$,
we impose a texture loss
%
%
\begin{equation}\label{eq:lossTexture}
\mathcal{L}_\mathrm{texture}(\mathcal{T}_\mathrm{dyn})
=
\sum_{\mathbf{u} \in \mathcal{R}}
\lvert
	\hat{\mathcal{F}}_{c',\mathbf{u}}
	(\Phi_{c',\mathbf{u}}(\mathbf{V}_\mathrm{fi}, \mathcal{T}_\mathrm{dyn}, \mathbf{I}_\mathrm{sh})
	-
	\mathcal{I}_{c',\mathbf{u}})
\rvert
\end{equation}
%
%
which renders our character using the dynamic texture regressed from TexNet and the geometry from our EgNet and DefNet and compares it to the real image.
Here, $\hat{\mathcal{F}}_{c,\mathbf{u}}$ is the eroded image foreground mask.
We apply an erosion to avoid that background pixels are projected into the dynamic texture if the predicted geometry does not perfectly align with the image.
$\mathbf{I}_\mathrm{sh}$ denotes the identity lighting.
In contrast to the previous rendering losses, we only supervise the network on the conditioning view and not on all camera views.
%
%
%
\subsection{Implementation Details}
\label{sec:implementation}
In all experiments, we use the Adam optimizer~\cite{kingma14}.
Due to the memory limits and training time, we randomly sample 40 cameras views (if available) for all multi-view losses.
The distance transform images have a resolution of $350 \times 350$.
The rendering resolution of the differentiable renderer is $512 \times 512$ ($643 \times 470$) for the training of DeltaNet and the lighting optimization and $1024 \times 1024$ ($1285 \times 940$) for the training of TexNet.
We train EGNet for $360{,}000$ iterations with a batch size of 40 where we balance the silhouette and ARAP term with $100.0$ and $1500.0$, respectively and used a learning rate of $0.0001$. 
This step takes 20 hours using 4 NVIDIA Quadro RTX 8000 with 48GB of memory.
The lighting is optimized with a batch size of 4, a learning rate of $0.0001$, and $30{,}000$ iterations.
This takes around 7 hours.
For training DeltaNet, we balanced the chroma, silhouette, and Laplacian loss with $0.03775$, $500.0$, and $100{,}000.0$, respectively.
Again, we train for $360{,}000$ iterations using a batch size of 8 and a learning rate of $0.0001$ which takes 2 days.
Finally, for training TexNet we use a batch size of 12 and a learning rate of $0.0001$.
We iterate $720{,}000$ times which takes 4 days.
%
%

%
%
\section{Results}
\label{sec:results}
All our results are computed on a machine with an AMD EPYC 7502P processing unit and a Nvidia Quadro RTX 8000 graphics card. 
Our approach can run at 38 frames per second (fps) at inference time and therefore allows interactive applications as discussed later.
For the first frame of a test sequence, we copy over the pose of the first frame as the "previous frames" of the motion window as there are no real previous frames.
%
%
%
\subsection{Dataset}
\label{sec:datasets}
\begin{figure*}[t]
	\centering
	\includegraphics[width=0.70\textwidth]{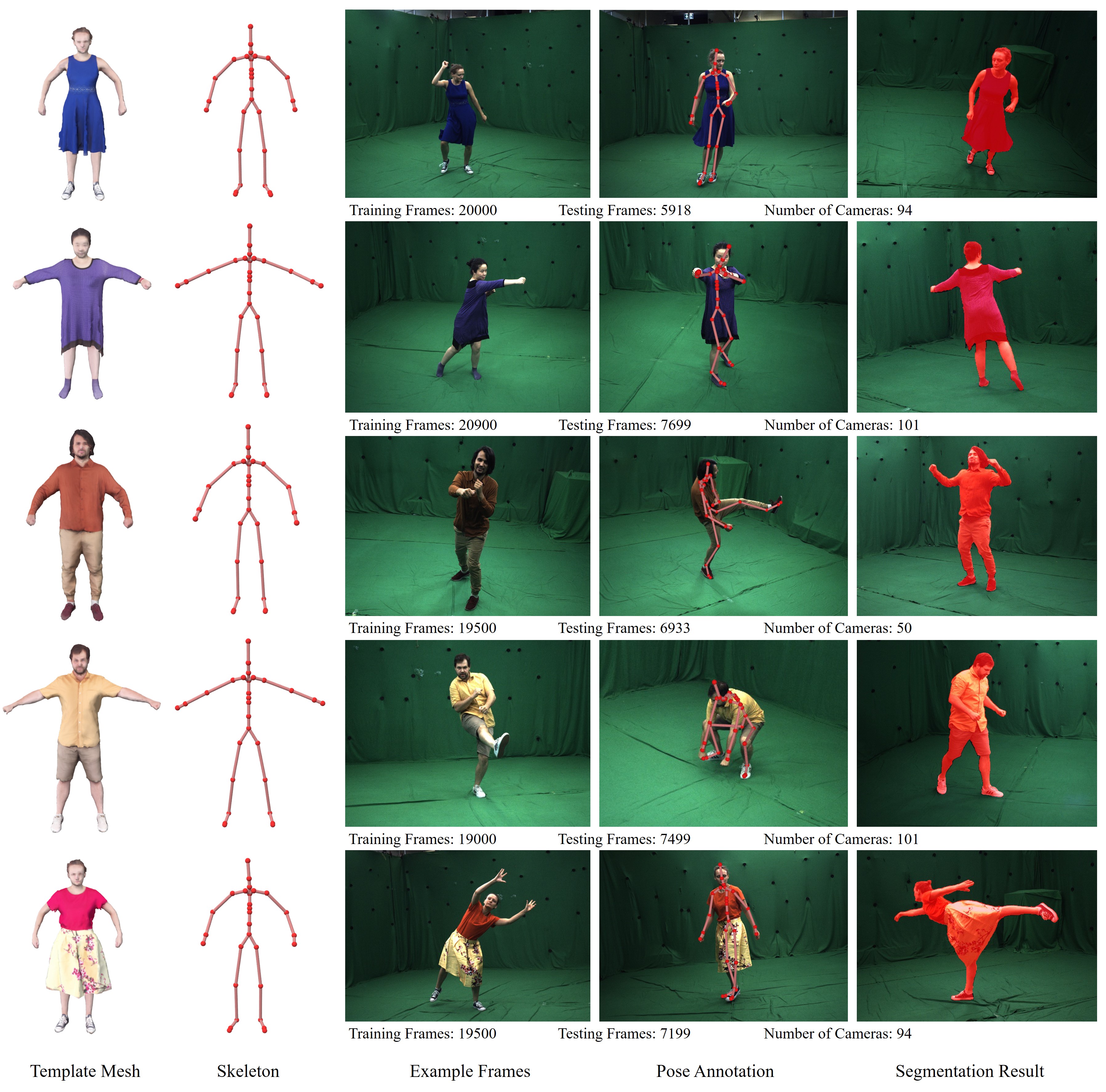} 
	\caption
	{
	DynaCap dataset. 
	We recorded 5 subjects wearing different types of apparel with multiple calibrated and synchronized cameras.
	Further, we capture a 3D template mesh and rig it to a skeleton.
	For each frame, we compute the ground truth 3D skeletal pose as well as ground truth foreground segmentation.
	}
	\label{fig:dataset}
\end{figure*}

%
%
We created a new dataset, called \textit{DynaCap}, which consists of 5 sequences containing 4 subjects wearing 5 different types of apparel, e.g., trousers and skirts (see Fig.~\ref{fig:dataset}).
Each sequence is recorded at 25fps and is split into a training and testing recording which contain around $20{,}000$ and $7000$ frames, respectively.
The training and test motions are significantly different from each other.
Following common practice, we acquired separate recordings for training and testing (instead of randomly sampling from a single sequence). 
For each sequence, we asked the subject to perform a wide range of motions like “dancing” which was freely interpreted by the subject.
We recorded with 50 to 101 synchronized and calibrated cameras at a resolution of $1285\times940$.
Further, we scanned each person to acquire a 3D template, as described in Sec.~\ref{sec:characterModel}, which is rigged to a skeleton.
For all sequences, we estimated the skeletal motion using \cite{captury} and segmented the foreground using color keying.
We will release the new dataset, as there are no other datasets available that target exactly such a setting, namely a single actor captured for a large range of motions and with such a dense camera setup.
Our dataset can be particularly interesting for dynamic neural scene representation approaches and can serve as a benchmark.
\par 
In addition, we use the subjects \textit{S1}, \textit{S2}, and \textit{S4} of the publicly available DeepCap dataset~\cite{habermann20} who wear trousers, T-shirts, skirts, and sleeves to evaluate our method also on external data which has a sparser camera setup.
The dataset comes along with ground truth pose tracking, calibrated, segmented, and synchronized multi-view imagery, in addition to a rigged template mesh.
Their dataset contains between 11 and 14 camera views at a resolution of $1024 \times 1024$ and a frame rate of 50fps.
%
%
%
%
%
%
\subsection{Qualitative Results}
\label{sec:qualiresults}
\begin{figure*}
	\centering
	\includegraphics[width=0.75\textwidth]{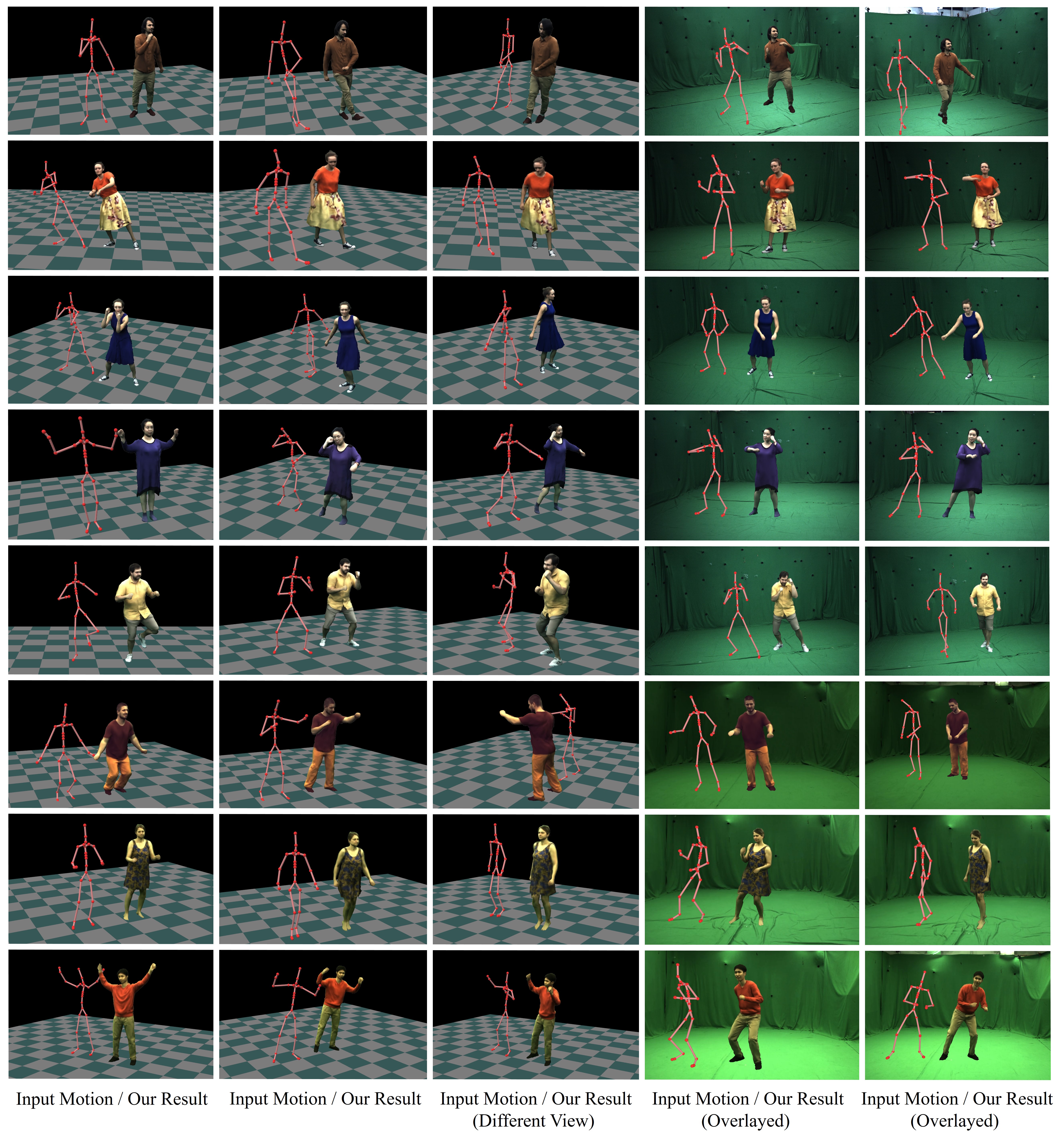} 
	\caption
	{
	Qualitative results. 
	From left to right. 
	Input testing pose and our posed, deformed and textured result shown from an arbitrary viewpoint.
	Our result for another testing pose and viewpoint. 
	The same pose as in second column but rendered from a different view point.
	Note the view-dependent appearance change on the skin and clothing due to view-dependent reflections.
	The last two columns show testing poses but viewed from the training camera viewpoints.
	This allows augmenting the empty background that we captured for each camera with our result.
	}
	\label{fig:qualitative}
	\vspace{-8pt}
\end{figure*}

In Fig.~\ref{fig:qualitative}, we illustrate results for all 8 sequences showing different types of apparel.
Again, note that our method learns videorealistic motion- and view-dependent dynamic surface deformation, including also deformations of loose apparel (such as the skirt and dress in Fig.~\ref{fig:qualitative}), without requiring a physics simulation, and texture \textit{only} from multi-view imagery and does \textit{not} require any dense 3D data such as depth maps, point clouds or registered meshes for supervision.
Our approach works not only well for tighter clothes such as pants but also for more dynamic ones like skirts.
We demonstrate that our approach can create videorealistic results for \textit{unseen} very challenging and fast motions, e.g., jumping jacks (see also supplemental video).
Moreover, the texture is consistent while changing the viewpoint (images without green screen background), which shows that our view conditioned TexNet also generalizes to novel view points.
The generalization comes from the fact that our networks for deformation regression as well as for texture regression focus on local configurations rather than the full 3D body motion. 
However, the network still allows global reasoning but this effect is dampened by the network design. 
Technically, this is accomplished by the local graph/image features and the graph/image convolutions.
Further, note the view-dependent effects like reflections on the skin and clothing (second and third column where we keep the pose fixed and only change the view point).
Given an image of the empty capture scene (images with green screen background), our approach allows augmenting the empty background image with our results to produce realistic looking images.
%
%
\par
\begin{figure}
	\centering
	\includegraphics[width=0.9\linewidth]{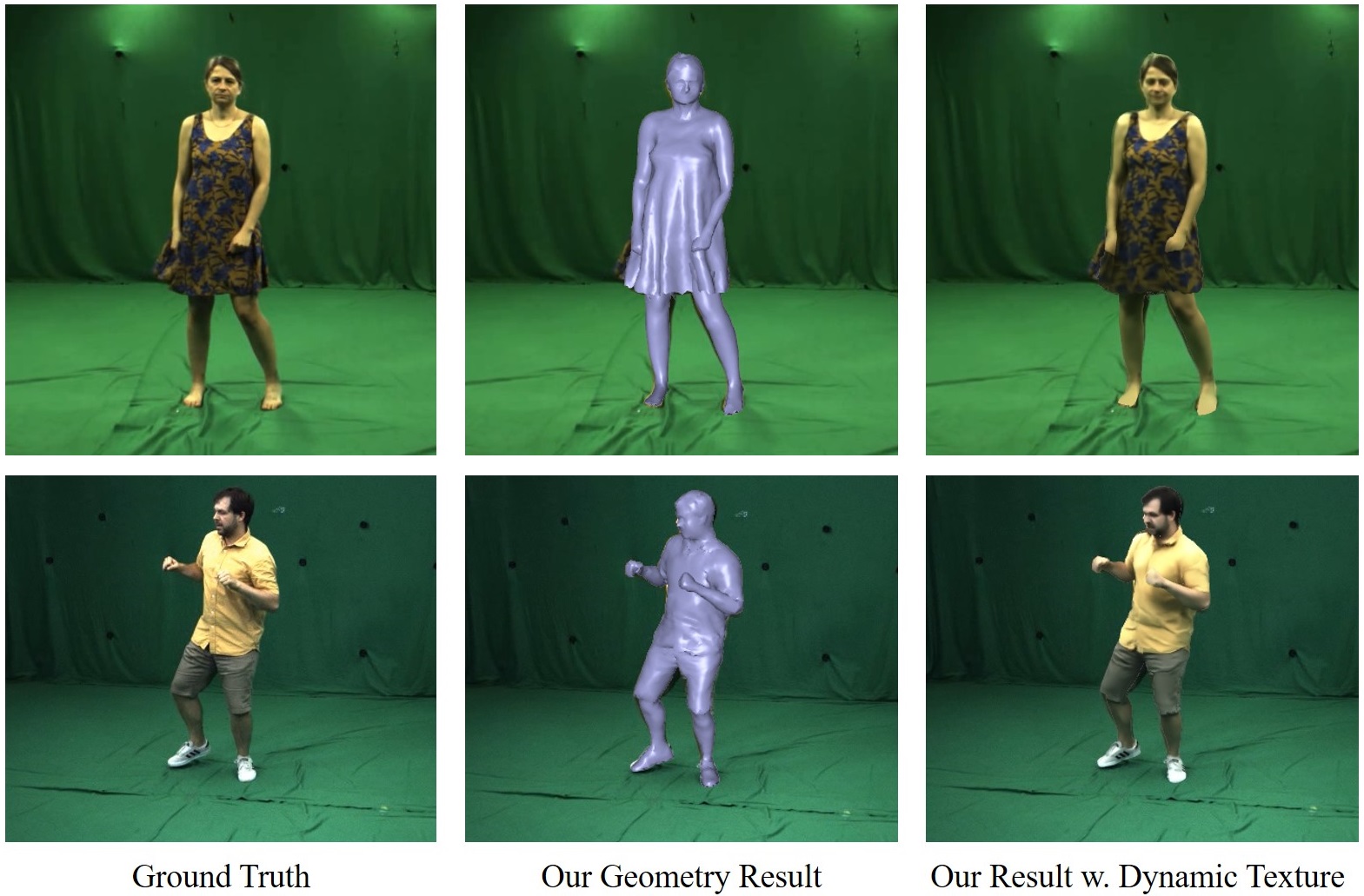} 
	\caption
	{
	Our geometry and texture networks generalize well to unseen motions as the geometry overlays nicely onto the ground truth frame and the final textured result looks almost identical to the ground truth.
	Importantly, our method does \textit{not} take the ground truth frame as an input as it only takes the unseen motion from the video.
	}
	\label{fig:qualitative1}
	\vspace{-12pt}
\end{figure}

Fig.~\ref{fig:qualitative1} shows that our predicted geometry (on test data) precisely overlays to the corresponding image which demonstrates that our approach generalizes well to unseen motions.
Importantly, the ground truth frame showing the actor is \textit{not} an input to our method as our method only takes the skeletal motion which we extracted from the video.
We also show our textured result overlayed onto the ground truth. 
Our TexNet generalizes well to unseen motions, captures the motion-dependent dynamics and looks photo-realistic as ground truth and our rendered result look almost identical.
%
%
%
%
%
%
%
%
%
\subsection{Comparison}
\label{sec:comparison}
Only few people in the research community have targeted creating video realistic characters from multi-view video that can be controlled to perform unseen skeleton motions. 
To the best of our knowledge, there are only three previous works~\cite{Xu:SIGGRPAH:2011, casas14, shysheya19} that also assume multi-view video data for building a controllable and textured character.
However, these works do not provide their code and thus are hard to compare to.
Moreover, they either do not share their data~\cite{shysheya19} or the publicly available sequences are too short for training our approach~\cite{Xu:SIGGRPAH:2011,casas14} and as well lack a textured template which our method assumes as given.
Therefore in Tab.~\ref{tab:comparison}, we resort to a conceptual comparison showing the advantage of our method as well as an extensive visual comparison in the supplemental video where we recreate similar scenes.
%
%
\par
The earlier works of \cite{Xu:SIGGRPAH:2011} and \cite{casas14} are both non learning-based and instead use texture retrieval to synthesize dynamic textures.
In contrast to our approach, they both suffer from the fact that their geometry is either fully driven by skinning based deformations~\cite{Xu:SIGGRPAH:2011} or by motion graphs~\cite{casas14}.
Thus, they cannot model motion-dependent geometric deformations and fail to model plausible dynamics of loose apparel, as our method can do it.
Moreover, as they rely on retrieval-based techniques, their approaches do not generalize well to motions different from motions in the dataset. 
Furthermore, the retrieval is expensive to compute, making real time application impossible.
In contrast, our approach leverages dedicated geometry networks (EGNet and DeltaNet) which predict motion-dependent geometry deformations for both tight and loose apparel.
Further, our approach enables animation and control in real-time, and generalizes well to unseen motions (see supplemental video and Fig.~\ref{fig:qualitative}).
%
%
\par
More recently, Textured Neural avatars~\cite{shysheya19} was proposed as the first learning based approach for creating controllable and textured characters using multi-view data.
In contrast to our approach, they do not model geometry explicitly but use DensePose~\cite{guel18} as a geometric proxy in image space.
As a consequence, their approach does not provide space-time coherent geometry as well as motion-dependent surface deformation which is important in most graphics and animation settings.
Moreover, they recover a static texture during training which prevents modelling motion- and view-dependent effects.
%
%
\par
Further, our supplemental video shows that our approach is a significant step forward in terms of visual quality compared to previous methods, as they either suffer from temporal inconsistency, sudden jumps in the texture originating from the texture retrieval, and missing body parts as geometry is not modelled explicitly.
%
%
\begin{table*}
	\begin{center}
		\caption
    	{
    	Conceptual comparison to previous multi-view based approaches for controllable character animation / synthesis~\cite{Xu:SIGGRPAH:2011, casas14, shysheya19}.
		Note that all previous works fall short in multiple desirable categories while our proposed approach fulfills all these requirements.
    	}
    	\label{tab:comparison}
		\begin{tabular}{|c|c|c|c|c|c|c|c|}
			\hline
			\multicolumn{8}{|c|}{\textit{Comparison to Previous Multi-view Based Methods}} \\
			\hline
			                             & Dyn. Geo.  & Dyn. Tex.   & View Dep. Effects    & Controllable  & Real-time  & Unseen Motions & Loose Clothing\\
			\hline
			\cite{Xu:SIGGRPAH:2011}      & \xmark	& \cmark	    & \cmark	    & \cmark	& \xmark	& \xmark	 & \xmark		    \\
			\cite{casas14}  			 & \xmark	& \cmark	    & \cmark	    & \cmark	& \xmark	& \xmark	 & \xmark		    \\
			\cite{shysheya19}  	         & \xmark	& \xmark	    & \xmark	    & \cmark    & \cmark	& \cmark     & \xmark			\\
			\textbf{Ours}  				 & \cmark   & \cmark	    & \cmark	    & \cmark	& \cmark	& \cmark	 & \cmark	        \\
			\hline
		\end{tabular}
	\end{center}
	\vspace{-12pt}
\end{table*}
%
%
%
%
\subsection{Quantitative Evaluation}
\label{sec:quantitative}
%
%
\subsubsection{Geometry}
\label{sec:recComparison}
To evaluate our approach in terms of geometry, we leverage the challenging \textit{S4} testing sequence ($11{,}000$ frames) of the DeepCap dataset~\cite{habermann20} shown in the top row of Fig.~\ref{fig:qualitative1}.
We trained our model on the corresponding multi-view training sequence and used their mesh template.
We follow the evaluation procedure described in the original paper.
Therefore, we measure the multi-view foreground mask overlap between ground truth foreground segmentation and the foreground mask obtained from our projected and deformed model on all available views (AMVIoU) averaged over every 100th frame.
%
%
\par
In Tab.~\ref{tab:reconstruction}, we compare to the multi-view baseline implementation of \cite{habermann20}, referred to as MVBL. 
Here, they perform optimization-based multi-view pose and surface fitting using sparse and dense image cues, e.g., 2D joint predictions and the foreground masks.
Importantly, they apply this on the \textit{testing} sequence directly whereas our method only takes the skeletal motion without even seeing the multi-view imagery.
Nonetheless, our results are more accurate than MVBL.
We found that their sequential optimization of pose and deformation can fall into erroneous local optima, resulting in worse overlay.
In contrast, our method benefits from the randomness of the stochastic gradient descent and the shuffling of data which reduces the likelihood of getting stuck in local optima.
We also compared our poseable and dynamic representation to the classical Dual Quaternion character skinning~\cite{kavan07} where we use the same poses as used for our approach to animate the rigged character.
Skinning can merely approximate skeleton induced surface deformation, but it fails to represent dynamic clothing deformations, as we can handle them.
Thus, we clearly outperform their approach as they cannot account for the surface deformation caused by the motion of the actor, e.g. swinging of a skirt.
%
%
\par
We report the same metrics also on the training data. 
Even from a reconstruction perspective our method produces accurate results during training and the proposed representation is able to fit the image observations almost perfectly.
Notably, there is only a small accuracy difference between training and testing performance. 
This confirms that our approach generalizes well to unseen motions.
%
%
\begin{table}
	\begin{center}
    	\caption
    	{
    		Accuracy of the surface deformation.
    		Note that we outperform the pure skinning based approach~\cite{kavan07} as they cannot account for dynamic cloth deformations.
    		Our method further improves over MVBL even though this optimization based approach sees the multi-view test images.
    		Finally, our approach performs similarly on training and testing data showing that the geometry networks generalize to unseen motions.
    	}
		\label{tab:reconstruction}
		\begin{tabular}{|c|c|}
			\hline
			\multicolumn{2}{|c|}{\textit{AMVIoU (in \%) on S4 sequence}} \\
			\hline
			\textbf{Method}                                 		        & \textbf{AMVIoU}$\uparrow$               		   	\\
			\hline
			MVBL~\cite{habermann20}                                         & 88.14		                                        \\
			\cite{kavan07}			                                        & 79.45		                                        \\
			\textbf{Ours} 							                        & \textbf{90.70}		                            \\
			\hline
		    Ours (Train)							                        & 94.07	                                            \\
			\hline
		\end{tabular}
	\end{center}
	\vspace{-12pt}
\end{table}
%
%
%
\subsubsection{Texture}
In Tab.~\ref{tab:texture}, we evaluate the realism of our motion-dependent dynamic texture on the same sequence as before (testing sequence of \textit{S4}).
We again trained on the training motion sequence of \textit{S4} but hold out the camera 4 as a test view.
We evaluate our approach on \textit{Train Camera} 0 and \textit{Test Camera} 4 for \textit{Train Motions} and \textit{Test Motions}.
Therefore, we compute the mean squared image error (MSE) and the structural similarity index measure (SSIM) between the rendered model and the ground truth multi-view images averaged over every 100th frame where we masked out the background as our approach does not synthesize the background.
Our method produces visually plausible results for novel motions rendered from a training view (see top row of Fig.~\ref{fig:qualitative1} and the $4th$ and $5th$ column of the second last row of Fig.~\ref{fig:qualitative}).
But also for novel motions \textit{and} novel camera views our approach produces video-realistic results (see $1th$, $2th$, and $3th$ column of the second last row of Fig.~\ref{fig:qualitative}).
Tab.~\ref{tab:texture} also quantitatively confirms this since all configurations of training/testing poses and camera views have a low MSE value and a high SSIM value.
While there is an accuracy drop between test and train, visually the quality only decreases slightly and the absolute accuracy for each configuration is comparably high.
%
%
\begin{table}
	\begin{center}
    	\caption
    	{
    		Photometric error in terms of MSE and SSIM averaged over every 100th frame.
    		Note that our approach achieves overall low MSE results and high SSIM values.
    		While the accuracy differs between test and train, the absolute accuracy is still comparably high and the visual quality only decreases slightly proving the generalization ability of our approach.
    	}
    	\label{tab:texture}	
		\begin{tabular}{|c|c|c|}
			\hline
			\multicolumn{3}{|c|}{\textit{Photometric Error} on \textit{S4}} \\
			\hline
			\textbf{Method}                                 		& \textbf{MSE} $\downarrow$   	            & \textbf{SSIM}$\uparrow$       \\
			\hline
		    Ours (Train Motion / Train Camera)							            & 14.79		                                        & 0.99054                             \\ 
		    Ours (Train Motion / Test Camera)							                & 31.44		                                        & 0.98610                         \\ 
		    Ours (Test Motion / Train Camera)							                & 29.00		                                        & 0.98357                         \\ 
			Ours (Test Motion / Test Camera)							                & 43.29		                                        & 0.98278                         \\ 
			\hline
		\end{tabular}
	\end{center}
	\vspace{-12pt}
\end{table}
%
%
%
%
\subsection{Ablation}
\label{sec:ablation}
%
%
\subsubsection{Deformation Modules}
%
%
First, we evaluate the design choices for predicting the surface deformations.
Therefore, we compare the impact of the DeltaNet against only using EGNet, which we refer to as \textit{EGNet-only}.
Tab.~\ref{tab:ablation1} clearly shows that the additional vertex displacements improve the reconstruction accuracy as they are able to capture finer wrinkles and deformations (see also Fig.~\ref{fig:withWithoutDelta}).
While the regressed embedded deformation still performs better than a pure skinning approach, it cannot completely match the ground truth silhouettes due to the limited graph resolution causing the slightly lower accuracy compared to using the displacements.
%
%
%
\par
We further evaluate the impact of our SAGC over two baselines using a fully connected (FC) architecture and an unstructured graph convolutional operator~\cite{defferrard17} where the latter is integrated into our overall architecture and therefore just replaces the SAGC operators.
We replace EGNet and DeltaNet with two fully connected networks that take the normalized motion angles as input, apply 19 fully connected layers (same depth as the proposed architecture) with nonlinear Elu activation functions, and output graph parameters and vertex displacements, respectively.
As the fully connected networks have no notion of locality, they are not able to generalize well.
Further, one can see that our proposed graph convolutional operation performs better than the one proposed by \cite{defferrard17} because the latter shares weights across nodes while we use node specific weights which are able to encode the underlying knowledge about the individual deformation behaviours of the surface.
%
%
%
\par
We also evaluate the importance of predicting the per-vertex displacements in the canonical pose space compared to predicting them in the global pose space.
Note that the disentanglement of pose and deformation helps with the generalization of the network which leads to better accuracy in terms of foreground overlay.
%
%
%
\par
\begin{figure}[t]
	\centering
	\includegraphics[width=\linewidth]{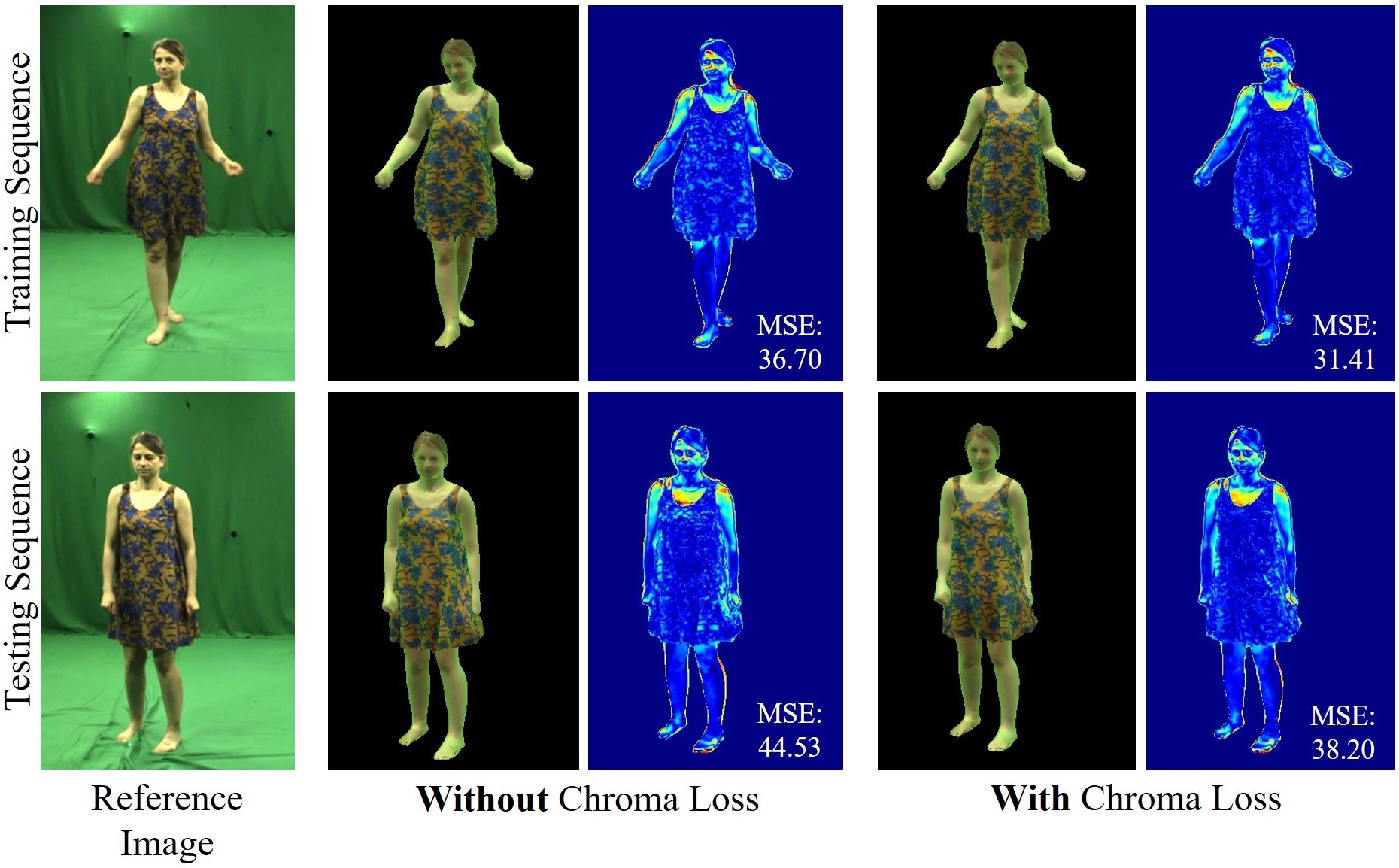} 
	\caption
	{
	   Impact of the chroma loss.
	   During training and testing, the chroma loss disambiguates drifts on the visual hull and gives more accurate results.
	}
	\label{fig:ablationChromaLoss}
	\vspace{-12pt}
\end{figure}

Finally, we evaluate the impact of the chroma loss compared to only using silhouette supervision.
Note that reporting the IoU would not be meaningful as the silhouette loss alone can already ensure matching silhouettes. 
However, drifts along the visual hull, carved by the silhouette images, cannot be well tracked by the silhouette term alone as shown in Fig.~\ref{fig:ablationChromaLoss}.
Our chroma loss penalizes these drifts, both, during training and testing leading to better results.
This can be best evaluated by comparing the MSE of the deformed model with the static texture and the ground truth image as shown in Fig.~\ref{fig:ablationChromaLoss}
Here, using the chroma loss has an error of 38.20 compared to an error of 44.53 when only the silhouette loss is used during test time.
This clearly shows that the chroma error can disambiguate drifts on the visual hull and thus gives more accurate results.
%
%
%
%
%
\subsubsection{Texture Module}
Next, we compare using our motion- and view-dependent texture to using a static texture rendered with and without optimized lighting.
Using the static texture without optimized lighting leads to the highest error. 
Optimizing the light already brings the rendering globally a bit closer to the ground truth image, but still fails to represent important dynamic and view-dependent effects.
By applying our dynamic texture also motion- and view-dependent texture effects can be captured, resulting in the lowest error.
%
%
\begin{table}[t]
	\begin{center}
	    \caption
    	{
    	Ablation study.
    	We evaluate the design choices for the geometry networks and texture networks.
    	Note that we beat the baselines in all aspects confirming that our design choices indeed lead to an improvement.
    	}
    	\label{tab:ablation1}
		\begin{tabular}{|c|c|c|}
			\hline
			\multicolumn{3}{|c|}{\textit{Ablation on the \textit{S4} sequence}} \\
			\hline
			\textbf{Method}                                        	& \textbf{AMVIoU}$\uparrow$  & \textbf{MSE}$\downarrow$  	\\
			\hline
			EGNet-only                                              & 87.89	            		& ---		   					\\
			Fully Connected                                         & 87.48	            		& ---		   					\\
			Unstructured GraphConv                                  & 83.30	            		& ---		   					\\
			Global Pose Space                                   & 89.81	        & ---                           \\
			\hline
			Without Lighting and Dynamic Texture                    & ---	            		& 176.99		   				\\
			Without Dynamic Texture                                 & ---            	    	& 60.50		   					\\
			\hline
			\textbf{Ours}    				  			            & \textbf{90.70}		   	& \textbf{43.29}				\\
			\hline
		\end{tabular}
	\end{center}
\end{table}
%
%
%
\subsubsection{Amount of Data}
Finally, we evaluate the influence of the number of training cameras for the \textit{OlekDesert} sequence in Tab.~\ref{tab:ablation2}.
We tested training with 5, 10, 25 and 49 cameras placed around the scene in a dome-like arrangement. 
We used the respective test motions for all reported metrics.
For computing the MSE, we chose camera 46 which was not part of the training views for all experiments.
Note that already 5 cameras can lead to plausible results.
Interestingly, with such a sparse setup our approach still produces coherent results for unseen viewpoints as the prediction is in canonical texture space, which implicitly regularizes the predictions, leading to a better generalization ability.
However, adding more cameras further improves both geometry and texture quality.
%
%
\begin{table}[t]
	\begin{center}
    	\caption
    	{
    	Influence of the number of available training cameras.
    	Already with few cameras our method achieves plausible results.
    	However, adding more cameras further improves the quality of both geometry and texture.
    	}
    	\label{tab:ablation2}
		\begin{tabular}{|c|c|c|}
			\hline
			\multicolumn{3}{|c|}{\textit{Ablation on the \textit{OlekDesert} sequence}} \\
			\hline
			\textbf{Method}                             &    \textbf{AMVIoU}$\uparrow$         	& \textbf{MSE}$\downarrow$  	\\
			\hline 
			5 camera views 								&	90.27           & 20.85		       			    	\\
			10 camera views 							&	90.34           & 19.32	           		 			\\
			25 camera views 							&	90.36	        & 17.49		           				\\
			\hline
		    \textbf{Ours (49 views)}				  	&	\textbf{90.68}	& \textbf{16.72}	       			\\
			\hline
		\end{tabular}
	\end{center}
	\vspace{-12pt}
\end{table}
%
%
%
%
\subsection{Applications}
\label{sec:applications}
\begin{figure}[t]
	\centering
	\includegraphics[width=0.9\linewidth]{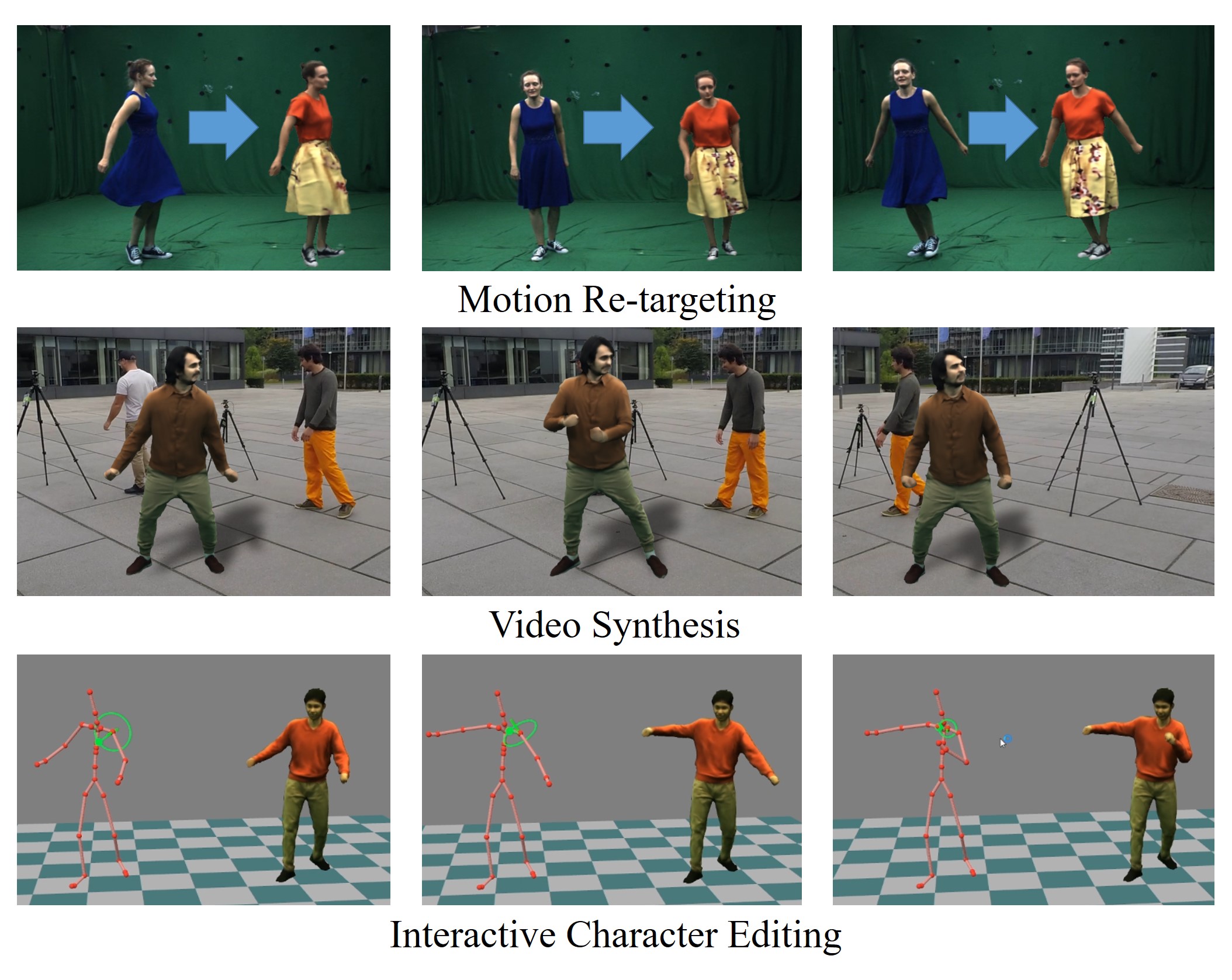} 
	\caption
	{
	Applications.
	Our method can be used in several applications such as motion re-targeting, neural video synthesis, and interactive character editing.
	Note that for all these applications, our method produces videorealistic results creating an immersive experience.
	}
	\label{fig:applications}
	\vspace{-12pt}
\end{figure}

As shown in Fig.~\ref{fig:applications}, our method can be used in several applications such as motion re-targeting where a source actor (blue dress girl) drives our character model (red shirt girl).
Further, our method synthesizes new free-viewpoint videos of an actor only with a driving motion sequence. 
Moreover, we implemented an interactive interface, where the user can freely change the skeletal pose and 3D camera viewpoint and our method produces the posed, deformed, and texture geometry in real time.
%
%
%
%

%
\subsection{Limitations}
%
%
Our approach approximates clothing dynamics in a data-driven and plausible way, but actual physics-based clothing animation may still lead to further improved results.
In future research, this could be handled by employing those physics-based priors in the learning process or even at inference.
Further, our method cannot handle apparent topological changes such as taking off pieces of apparel.
We believe the current progress in implicit representations combined with our representation could help to generate such changes even though they are radically different from the initial template mesh.
We do not track the facial expression and hands.
2D face landmark trackers as well as hand trackers could be used to also track hands and face so that they can also be controlled in the deep dynamic character.
Currently, the training time of the network modules is quite long.
In the future, more efficient training schemes could be explored to solve this issue.
Moreover, we rely on good foreground segmentation results.
In consequence, our method might receive a wrong silhouette supervision when multiple people or other moving objects, which are detected as foreground, are in the training scene.
Explicitly modeling multi-person scenes and using a learning based multi-person detector could help here.
Finally, severely articulated poses like a hand stand, which are not within the training motion distribution, can lead to wrong deformation and texture predictions.
\section{Conclusion}
%
%
\par
We presented a real-time method that allows to animate the dynamic 3D surface deformation and texture of highly realistic 3D avatars in a user-controllable way.
Skeleton motion can be freely controlled and avatars can be free-viewpoint rendered from any 3D viewpoint.
To this end, we propose a learning based architecture which not only regresses dynamic surface deformations but also dynamic textures.
Our approach does not require any ground truth 3D supervision.
Instead, we only need multi-view imagery and employ new analysis-by-synthesis losses for supervision.
Our results outperform the state of the art in terms of surface detail and textural appearance and therefore the high visual quality of our animations opens up new possibilities in video-realistic character animation, controllable free-viewpoint video, and neural video synthesis.

\begin{acks}
All data captures and evaluations were performed at MPII by MPII.
The authors from MPII were supported by the ERC Consolidator Grant 4DRepLy (770784), the Deutsche Forschungsgemeinschaft (Project Nr. 409792180, Emmy Noether Programme, project: Real Virtual Humans) and Lise Meitner Postdoctoral Fellowship.
\end{acks}
%

\typeout{} 
\bibliographystyle{ACM-Reference-Format}
\bibliography{bib}


\begin{thebibliography}{101}


\ifx \showCODEN    \undefined \def \showCODEN     #1{\unskip}     \fi
\ifx \showDOI      \undefined \def \showDOI       #1{#1}\fi
\ifx \showISBNx    \undefined \def \showISBNx     #1{\unskip}     \fi
\ifx \showISBNxiii \undefined \def \showISBNxiii  #1{\unskip}     \fi
\ifx \showISSN     \undefined \def \showISSN      #1{\unskip}     \fi
\ifx \showLCCN     \undefined \def \showLCCN      #1{\unskip}     \fi
\ifx \shownote     \undefined \def \shownote      #1{#1}          \fi
\ifx \showarticletitle \undefined \def \showarticletitle #1{#1}   \fi
\ifx \showURL      \undefined \def \showURL       {\relax}        \fi
\providecommand\bibfield[2]{#2}
\providecommand\bibinfo[2]{#2}
\providecommand\natexlab[1]{#1}
\providecommand\showeprint[2][]{arXiv:#2}

\bibitem[\protect\citeauthoryear{Aberman, Shi, Liao, Lischinski, Chen, and
  Cohen{-}Or}{Aberman et~al\mbox{.}}{2019}]%
        {Lischinski2018}
\bibfield{author}{\bibinfo{person}{Kfir Aberman}, \bibinfo{person}{Mingyi Shi},
  \bibinfo{person}{Jing Liao}, \bibinfo{person}{Dani Lischinski},
  \bibinfo{person}{Baoquan Chen}, {and} \bibinfo{person}{Daniel Cohen{-}Or}.}
  \bibinfo{year}{2019}\natexlab{}.
\newblock \showarticletitle{Deep Video-Based Performance Cloning}.
\newblock \bibinfo{journal}{\emph{Comput. Graph. Forum}} \bibinfo{volume}{38},
  \bibinfo{number}{2} (\bibinfo{year}{2019}), \bibinfo{pages}{219--233}.
\newblock
\urldef\tempurl%
\url{https://doi.org/10.1111/cgf.13632}
\showDOI{\tempurl}


\bibitem[\protect\citeauthoryear{Alldieck, Magnor, Bhatnagar, Theobalt, and
  Pons-Moll}{Alldieck et~al\mbox{.}}{2019}]%
        {alldieck19}
\bibfield{author}{\bibinfo{person}{Thiemo Alldieck}, \bibinfo{person}{Marcus
  Magnor}, \bibinfo{person}{Bharat~Lal Bhatnagar}, \bibinfo{person}{Christian
  Theobalt}, {and} \bibinfo{person}{Gerard Pons-Moll}.}
  \bibinfo{year}{2019}\natexlab{}.
\newblock \showarticletitle{Learning to Reconstruct People in Clothing from a
  Single {RGB} Camera}. In \bibinfo{booktitle}{\emph{{IEEE}/{CVF} Conference on
  Computer Vision and Pattern Recognition ({CVPR})}}.
  \bibinfo{pages}{1175--1186}.
\newblock


\bibitem[\protect\citeauthoryear{Alldieck, Magnor, Xu, Theobalt, and
  Pons-Moll}{Alldieck et~al\mbox{.}}{2018a}]%
        {alldieck18b}
\bibfield{author}{\bibinfo{person}{Thiemo Alldieck}, \bibinfo{person}{Marcus
  Magnor}, \bibinfo{person}{Weipeng Xu}, \bibinfo{person}{Christian Theobalt},
  {and} \bibinfo{person}{Gerard Pons-Moll}.} \bibinfo{year}{2018}\natexlab{a}.
\newblock \showarticletitle{Detailed Human Avatars from Monocular Video}. In
  \bibinfo{booktitle}{\emph{International Conference on 3D Vision}}.
  \bibinfo{pages}{98--109}.
\newblock
\urldef\tempurl%
\url{https://doi.org/10.1109/3{DV}.2018.00022}
\showDOI{\tempurl}


\bibitem[\protect\citeauthoryear{Alldieck, Magnor, Xu, Theobalt, and
  Pons-Moll}{Alldieck et~al\mbox{.}}{2018b}]%
        {alldieck18a}
\bibfield{author}{\bibinfo{person}{Thiemo Alldieck}, \bibinfo{person}{Marcus
  Magnor}, \bibinfo{person}{Weipeng Xu}, \bibinfo{person}{Christian Theobalt},
  {and} \bibinfo{person}{Gerard Pons-Moll}.} \bibinfo{year}{2018}\natexlab{b}.
\newblock \showarticletitle{Video Based Reconstruction of 3D People Models}. In
  \bibinfo{booktitle}{\emph{{IEEE} Conference on Computer Vision and Pattern
  Recognition}}.
\newblock
\newblock
\shownote{{CVPR} Spotlight Paper.}


\bibitem[\protect\citeauthoryear{Bailey, Otte, Dilorenzo, and O'Brien}{Bailey
  et~al\mbox{.}}{2018}]%
        {Bailey:2018:FDD}
\bibfield{author}{\bibinfo{person}{Stephen~W. Bailey}, \bibinfo{person}{Dave
  Otte}, \bibinfo{person}{Paul Dilorenzo}, {and} \bibinfo{person}{James~F.
  O'Brien}.} \bibinfo{year}{2018}\natexlab{}.
\newblock \showarticletitle{Fast and Deep Deformation Approximations}.
\newblock \bibinfo{journal}{\emph{ACM Transactions on Graphics}}
  \bibinfo{volume}{37}, \bibinfo{number}{4} (\bibinfo{date}{Aug.}
  \bibinfo{year}{2018}), \bibinfo{pages}{119:1--12}.
\newblock
\urldef\tempurl%
\url{https://doi.org/10.1145/3197517.3201300}
\showDOI{\tempurl}
\newblock
\shownote{Presented at SIGGRAPH 2018, Los Angeles.}


\bibitem[\protect\citeauthoryear{Bhatnagar, Tiwari, Theobalt, and
  Pons-Moll}{Bhatnagar et~al\mbox{.}}{2019}]%
        {bhatnagar19}
\bibfield{author}{\bibinfo{person}{Bharat~Lal Bhatnagar},
  \bibinfo{person}{Garvita Tiwari}, \bibinfo{person}{Christian Theobalt}, {and}
  \bibinfo{person}{Gerard Pons-Moll}.} \bibinfo{year}{2019}\natexlab{}.
\newblock \showarticletitle{Multi-Garment Net: Learning to Dress 3D People from
  Images}. In \bibinfo{booktitle}{\emph{{IEEE} International Conference on
  Computer Vision ({ICCV})}}. {IEEE}.
\newblock


\bibitem[\protect\citeauthoryear{{{Blender}}}{{{Blender}}}{2020}]%
        {blender}
{{Blender}} \bibinfo{year}{2020}\natexlab{}.
\newblock \bibinfo{title}{{Blender}}.
\newblock \bibinfo{howpublished}{\url{https://www.blender.org/}}.
\newblock


\bibitem[\protect\citeauthoryear{Borgefors}{Borgefors}{1986}]%
        {borgefors86}
\bibfield{author}{\bibinfo{person}{Gunilla Borgefors}.}
  \bibinfo{year}{1986}\natexlab{}.
\newblock \showarticletitle{Distance transformations in digital images}.
\newblock \bibinfo{journal}{\emph{Computer Vision, Graphics, and Image
  Processing}} \bibinfo{volume}{34}, \bibinfo{number}{3}
  (\bibinfo{year}{1986}), \bibinfo{pages}{344 -- 371}.
\newblock
\showISSN{0734-189X}
\urldef\tempurl%
\url{https://doi.org/10.1016/S0734-189X(86)80047-0}
\showDOI{\tempurl}


\bibitem[\protect\citeauthoryear{Carranza, Theobalt, Magnor, and
  Seidel}{Carranza et~al\mbox{.}}{2003}]%
        {carranza03}
\bibfield{author}{\bibinfo{person}{Joel Carranza}, \bibinfo{person}{Christian
  Theobalt}, \bibinfo{person}{Marcus~A. Magnor}, {and}
  \bibinfo{person}{Hans-Peter Seidel}.} \bibinfo{year}{2003}\natexlab{}.
\newblock \showarticletitle{Free-viewpoint Video of Human Actors}.
\newblock \bibinfo{journal}{\emph{ACM Trans. Graph.}} \bibinfo{volume}{22},
  \bibinfo{number}{3} (\bibinfo{date}{July} \bibinfo{year}{2003}).
\newblock


\bibitem[\protect\citeauthoryear{Casas, Volino, Collomosse, and Hilton}{Casas
  et~al\mbox{.}}{2014}]%
        {casas14}
\bibfield{author}{\bibinfo{person}{Dan Casas}, \bibinfo{person}{Marco Volino},
  \bibinfo{person}{John Collomosse}, {and} \bibinfo{person}{Adrian Hilton}.}
  \bibinfo{year}{2014}\natexlab{}.
\newblock \showarticletitle{4D Video Textures for Interactive Character
  Appearance}.
\newblock \bibinfo{journal}{\emph{Comput. Graph. Forum}} \bibinfo{volume}{33},
  \bibinfo{number}{2} (\bibinfo{date}{May} \bibinfo{year}{2014}),
  \bibinfo{pages}{371–380}.
\newblock
\showISSN{0167-7055}
\urldef\tempurl%
\url{https://doi.org/10.1111/cgf.12296}
\showDOI{\tempurl}


\bibitem[\protect\citeauthoryear{Chan, Ginosar, Zhou, and Efros}{Chan
  et~al\mbox{.}}{2019}]%
        {chan2019dance}
\bibfield{author}{\bibinfo{person}{Caroline Chan}, \bibinfo{person}{Shiry
  Ginosar}, \bibinfo{person}{Tinghui Zhou}, {and} \bibinfo{person}{Alexei~A
  Efros}.} \bibinfo{year}{2019}\natexlab{}.
\newblock \showarticletitle{Everybody Dance Now}. In
  \bibinfo{booktitle}{\emph{International Conference on Computer Vision
  (ICCV)}}.
\newblock


\bibitem[\protect\citeauthoryear{Chen, Gao, Ling, Smith, Lehtinen, Jacobson,
  and Fidler}{Chen et~al\mbox{.}}{2019}]%
        {chen2019dibrender}
\bibfield{author}{\bibinfo{person}{Wenzheng Chen}, \bibinfo{person}{Jun Gao},
  \bibinfo{person}{Huan Ling}, \bibinfo{person}{Edward Smith},
  \bibinfo{person}{Jaakko Lehtinen}, \bibinfo{person}{Alec Jacobson}, {and}
  \bibinfo{person}{Sanja Fidler}.} \bibinfo{year}{2019}\natexlab{}.
\newblock \showarticletitle{Learning to Predict 3D Objects with an
  Interpolation-based Differentiable Renderer}. In
  \bibinfo{booktitle}{\emph{Advances In Neural Information Processing
  Systems}}.
\newblock


\bibitem[\protect\citeauthoryear{Choi, Moon, and Lee}{Choi
  et~al\mbox{.}}{2020}]%
        {Choi_2020_ECCV_Pose2Mesh}
\bibfield{author}{\bibinfo{person}{Hongsuk Choi}, \bibinfo{person}{Gyeongsik
  Moon}, {and} \bibinfo{person}{Kyoung~Mu Lee}.}
  \bibinfo{year}{2020}\natexlab{}.
\newblock \showarticletitle{Pose2Mesh: Graph Convolutional Network for 3D Human
  Pose and Mesh Recovery from a 2D Human Pose}. In
  \bibinfo{booktitle}{\emph{European Conference on Computer Vision (ECCV)}}.
\newblock


\bibitem[\protect\citeauthoryear{Choi and Ko}{Choi and Ko}{2005}]%
        {Choi05}
\bibfield{author}{\bibinfo{person}{Kwang-Jin Choi} {and} \bibinfo{person}{H.
  Ko}.} \bibinfo{year}{2005}\natexlab{}.
\newblock \showarticletitle{Research problems in clothing simulation}.
\newblock \bibinfo{journal}{\emph{Comput. Aided Des.}}  \bibinfo{volume}{37}
  (\bibinfo{year}{2005}), \bibinfo{pages}{585--592}.
\newblock


\bibitem[\protect\citeauthoryear{Collet, Chuang, Sweeney, Gillett, Evseev,
  Calabrese, Hoppe, Kirk, and Sullivan}{Collet et~al\mbox{.}}{2015}]%
        {collet15}
\bibfield{author}{\bibinfo{person}{Alvaro Collet}, \bibinfo{person}{Ming
  Chuang}, \bibinfo{person}{Pat Sweeney}, \bibinfo{person}{Don Gillett},
  \bibinfo{person}{Dennis Evseev}, \bibinfo{person}{David Calabrese},
  \bibinfo{person}{Hugues Hoppe}, \bibinfo{person}{Adam Kirk}, {and}
  \bibinfo{person}{Steve Sullivan}.} \bibinfo{year}{2015}\natexlab{}.
\newblock \showarticletitle{High-quality streamable free-viewpoint video}.
\newblock \bibinfo{journal}{\emph{ACM Transactions on Graphics (TOG)}}
  \bibinfo{volume}{34}, \bibinfo{number}{4} (\bibinfo{year}{2015}),
  \bibinfo{pages}{69}.
\newblock


\bibitem[\protect\citeauthoryear{Defferrard, Bresson, and
  Vandergheynst}{Defferrard et~al\mbox{.}}{2017}]%
        {defferrard17}
\bibfield{author}{\bibinfo{person}{Michaël Defferrard},
  \bibinfo{person}{Xavier Bresson}, {and} \bibinfo{person}{Pierre
  Vandergheynst}.} \bibinfo{year}{2017}\natexlab{}.
\newblock \bibinfo{title}{Convolutional Neural Networks on Graphs with Fast
  Localized Spectral Filtering}.
\newblock
\newblock
\showeprint[arxiv]{1606.09375}~[cs.LG]


\bibitem[\protect\citeauthoryear{Esser, Haux, Milbich, and orn Ommer}{Esser
  et~al\mbox{.}}{2018}]%
        {Esser_2018_ECCV_Workshops}
\bibfield{author}{\bibinfo{person}{Patrick Esser}, \bibinfo{person}{Johannes
  Haux}, \bibinfo{person}{Timo Milbich}, {and} \bibinfo{person}{Bj orn Ommer}.}
  \bibinfo{year}{2018}\natexlab{}.
\newblock \showarticletitle{Towards Learning a Realistic Rendering of Human
  Behavior}. In \bibinfo{booktitle}{\emph{The European Conference on Computer
  Vision (ECCV) Workshops}}.
\newblock


\bibitem[\protect\citeauthoryear{Feng, Yu, and Kim}{Feng et~al\mbox{.}}{2010}]%
        {fengww10}
\bibfield{author}{\bibinfo{person}{Wei-Wen Feng}, \bibinfo{person}{Yizhou Yu},
  {and} \bibinfo{person}{Byung-Uck Kim}.} \bibinfo{year}{2010}\natexlab{}.
\newblock \showarticletitle{A Deformation Transformer for Real-Time Cloth
  Animation}.
\newblock \bibinfo{journal}{\emph{ACM Trans. Graph.}} \bibinfo{volume}{29},
  \bibinfo{number}{4}, Article \bibinfo{articleno}{108} (\bibinfo{date}{July}
  \bibinfo{year}{2010}), \bibinfo{numpages}{9}~pages.
\newblock
\showISSN{0730-0301}
\urldef\tempurl%
\url{https://doi.org/10.1145/1778765.1778845}
\showDOI{\tempurl}


\bibitem[\protect\citeauthoryear{Gafni, Thies, Zollhöfer, and Nießner}{Gafni
  et~al\mbox{.}}{2020}]%
        {gafni2020dynamic}
\bibfield{author}{\bibinfo{person}{Guy Gafni}, \bibinfo{person}{Justus Thies},
  \bibinfo{person}{Michael Zollhöfer}, {and} \bibinfo{person}{Matthias
  Nießner}.} \bibinfo{year}{2020}\natexlab{}.
\newblock \bibinfo{title}{Dynamic Neural Radiance Fields for Monocular 4D
  Facial Avatar Reconstruction}.
\newblock
\newblock
\showeprint[arxiv]{2012.03065}~[cs.CV]


\bibitem[\protect\citeauthoryear{Guan, Reiss, Hirshberg, Weiss, and Black}{Guan
  et~al\mbox{.}}{2012}]%
        {guan12}
\bibfield{author}{\bibinfo{person}{Peng Guan}, \bibinfo{person}{Loretta Reiss},
  \bibinfo{person}{David~A. Hirshberg}, \bibinfo{person}{Alexander Weiss},
  {and} \bibinfo{person}{Michael~J. Black}.} \bibinfo{year}{2012}\natexlab{}.
\newblock \showarticletitle{DRAPE: DRessing Any PErson}.
\newblock \bibinfo{journal}{\emph{ACM Trans. Graph.}} \bibinfo{volume}{31},
  \bibinfo{number}{4}, Article \bibinfo{articleno}{35} (\bibinfo{date}{July}
  \bibinfo{year}{2012}), \bibinfo{numpages}{10}~pages.
\newblock
\showISSN{0730-0301}
\urldef\tempurl%
\url{https://doi.org/10.1145/2185520.2185531}
\showDOI{\tempurl}


\bibitem[\protect\citeauthoryear{G{\"u}ler, Neverova, and Kokkinos}{G{\"u}ler
  et~al\mbox{.}}{2018}]%
        {guel18}
\bibfield{author}{\bibinfo{person}{Riza~Alp G{\"u}ler},
  \bibinfo{person}{Natalia Neverova}, {and} \bibinfo{person}{Iasonas
  Kokkinos}.} \bibinfo{year}{2018}\natexlab{}.
\newblock \showarticletitle{DensePose: Dense Human Pose Estimation In The
  Wild}. In \bibinfo{booktitle}{\emph{The IEEE Conference on Computer Vision
  and Pattern Recognition (CVPR)}}.
\newblock


\bibitem[\protect\citeauthoryear{Gundogdu, Constantin, Seifoddini, Dang,
  Salzmann, and Fua}{Gundogdu et~al\mbox{.}}{2019}]%
        {gundogdu19}
\bibfield{author}{\bibinfo{person}{Erhan Gundogdu}, \bibinfo{person}{Victor
  Constantin}, \bibinfo{person}{Amrollah Seifoddini}, \bibinfo{person}{Minh
  Dang}, \bibinfo{person}{Mathieu Salzmann}, {and} \bibinfo{person}{Pascal
  Fua}.} \bibinfo{year}{2019}\natexlab{}.
\newblock \showarticletitle{Garnet: A Two-stream Network for Fast and Accurate
  3D Cloth Draping}. In \bibinfo{booktitle}{\emph{{IEEE} International
  Conference on Computer Vision ({ICCV})}}. {IEEE}.
\newblock


\bibitem[\protect\citeauthoryear{Habermann, Xu, Zollhoefer, Pons-Moll, and
  Theobalt}{Habermann et~al\mbox{.}}{2019}]%
        {habermann19}
\bibfield{author}{\bibinfo{person}{Marc Habermann}, \bibinfo{person}{Weipeng
  Xu}, \bibinfo{person}{Michael Zollhoefer}, \bibinfo{person}{Gerard
  Pons-Moll}, {and} \bibinfo{person}{Christian Theobalt}.}
  \bibinfo{year}{2019}\natexlab{}.
\newblock \showarticletitle{LiveCap: Real-time Human Performance Capture from
  Monocular Video}.
\newblock \bibinfo{journal}{\emph{ACM Trans. Graph.}} (\bibinfo{year}{2019}).
\newblock


\bibitem[\protect\citeauthoryear{Habermann, Xu, Zollhoefer, Pons-Moll, and
  Theobalt}{Habermann et~al\mbox{.}}{2020}]%
        {habermann20}
\bibfield{author}{\bibinfo{person}{Marc Habermann}, \bibinfo{person}{Weipeng
  Xu}, \bibinfo{person}{Michael Zollhoefer}, \bibinfo{person}{Gerard
  Pons-Moll}, {and} \bibinfo{person}{Christian Theobalt}.}
  \bibinfo{year}{2020}\natexlab{}.
\newblock \showarticletitle{DeepCap: Monocular Human Performance Capture Using
  Weak Supervision}. In \bibinfo{booktitle}{\emph{{IEEE} Conference on Computer
  Vision and Pattern Recognition (CVPR)}}.
\newblock


\bibitem[\protect\citeauthoryear{Hahn, Thomaszewski, Coros, Sumner, Cole,
  Meyer, DeRose, and Gross}{Hahn et~al\mbox{.}}{2014}]%
        {Hahn14}
\bibfield{author}{\bibinfo{person}{Fabian Hahn}, \bibinfo{person}{Bernhard
  Thomaszewski}, \bibinfo{person}{Stelian Coros}, \bibinfo{person}{Robert~W.
  Sumner}, \bibinfo{person}{Forrester Cole}, \bibinfo{person}{Mark Meyer},
  \bibinfo{person}{Tony DeRose}, {and} \bibinfo{person}{Markus Gross}.}
  \bibinfo{year}{2014}\natexlab{}.
\newblock \showarticletitle{Subspace Clothing Simulation Using Adaptive Bases}.
\newblock \bibinfo{journal}{\emph{ACM Trans. Graph.}} \bibinfo{volume}{33},
  \bibinfo{number}{4}, Article \bibinfo{articleno}{105} (\bibinfo{date}{July}
  \bibinfo{year}{2014}), \bibinfo{numpages}{9}~pages.
\newblock
\showISSN{0730-0301}
\urldef\tempurl%
\url{https://doi.org/10.1145/2601097.2601160}
\showDOI{\tempurl}


\bibitem[\protect\citeauthoryear{He, Zhang, Ren, and Sun}{He
  et~al\mbox{.}}{2016}]%
        {he16}
\bibfield{author}{\bibinfo{person}{Kaiming He}, \bibinfo{person}{Xiangyu
  Zhang}, \bibinfo{person}{Shaoqing Ren}, {and} \bibinfo{person}{Jian Sun}.}
  \bibinfo{year}{2016}\natexlab{}.
\newblock \showarticletitle{Deep Residual Learning for Image Recognition}. In
  \bibinfo{booktitle}{\emph{IEEE Conference on Computer Vision and Pattern
  Recognition (CVPR)}}.
\newblock


\bibitem[\protect\citeauthoryear{Hilsmann, Fechteler, Morgenstern, Paier,
  Feldmann, Schreer, and Eisert}{Hilsmann et~al\mbox{.}}{2020}]%
        {hilsmann20}
\bibfield{author}{\bibinfo{person}{Anna Hilsmann}, \bibinfo{person}{Philipp
  Fechteler}, \bibinfo{person}{Wieland Morgenstern}, \bibinfo{person}{Wolfgang
  Paier}, \bibinfo{person}{Ingo Feldmann}, \bibinfo{person}{Oliver Schreer},
  {and} \bibinfo{person}{Peter Eisert}.} \bibinfo{year}{2020}\natexlab{}.
\newblock \showarticletitle{Going beyond free viewpoint: creating animatable
  volumetric video of human performances}.
\newblock \bibinfo{journal}{\emph{IET Computer Vision}} \bibinfo{volume}{14},
  \bibinfo{number}{6} (\bibinfo{date}{Sep} \bibinfo{year}{2020}),
  \bibinfo{pages}{350–358}.
\newblock
\showISSN{1751-9640}
\urldef\tempurl%
\url{https://doi.org/10.1049/iet-cvi.2019.0786}
\showDOI{\tempurl}


\bibitem[\protect\citeauthoryear{Holden, Komura, and Saito}{Holden
  et~al\mbox{.}}{2017}]%
        {10.1145/3072959.3073663}
\bibfield{author}{\bibinfo{person}{Daniel Holden}, \bibinfo{person}{Taku
  Komura}, {and} \bibinfo{person}{Jun Saito}.} \bibinfo{year}{2017}\natexlab{}.
\newblock \showarticletitle{Phase-Functioned Neural Networks for Character
  Control}.
\newblock \bibinfo{journal}{\emph{ACM Trans. Graph.}} \bibinfo{volume}{36},
  \bibinfo{number}{4}, Article \bibinfo{articleno}{42} (\bibinfo{date}{July}
  \bibinfo{year}{2017}), \bibinfo{numpages}{13}~pages.
\newblock
\showISSN{0730-0301}
\urldef\tempurl%
\url{https://doi.org/10.1145/3072959.3073663}
\showDOI{\tempurl}


\bibitem[\protect\citeauthoryear{Isola, Zhu, Zhou, and Efros}{Isola
  et~al\mbox{.}}{2016}]%
        {isola18}
\bibfield{author}{\bibinfo{person}{Phillip Isola}, \bibinfo{person}{Jun{-}Yan
  Zhu}, \bibinfo{person}{Tinghui Zhou}, {and} \bibinfo{person}{Alexei~A.
  Efros}.} \bibinfo{year}{2016}\natexlab{}.
\newblock \showarticletitle{Image-to-Image Translation with Conditional
  Adversarial Networks}.
\newblock \bibinfo{journal}{\emph{CoRR}}  \bibinfo{volume}{abs/1611.07004}
  (\bibinfo{year}{2016}).
\newblock
\showeprint[arxiv]{1611.07004}
\urldef\tempurl%
\url{http://arxiv.org/abs/1611.07004}
\showURL{%
\tempurl}


\bibitem[\protect\citeauthoryear{Isola, Zhu, Zhou, and Efros}{Isola
  et~al\mbox{.}}{2017}]%
        {pix2pix2017}
\bibfield{author}{\bibinfo{person}{Phillip Isola}, \bibinfo{person}{Jun-Yan
  Zhu}, \bibinfo{person}{Tinghui Zhou}, {and} \bibinfo{person}{Alexei~A
  Efros}.} \bibinfo{year}{2017}\natexlab{}.
\newblock \showarticletitle{Image-to-Image Translation with Conditional
  Adversarial Networks}.
\newblock \bibinfo{journal}{\emph{CVPR}} (\bibinfo{year}{2017}).
\newblock


\bibitem[\protect\citeauthoryear{Jiang, Ji, Han, and Zwicker}{Jiang
  et~al\mbox{.}}{2020}]%
        {jiang2020sdfdiff}
\bibfield{author}{\bibinfo{person}{Yue Jiang}, \bibinfo{person}{Dantong Ji},
  \bibinfo{person}{Zhizhong Han}, {and} \bibinfo{person}{Matthias Zwicker}.}
  \bibinfo{year}{2020}\natexlab{}.
\newblock \showarticletitle{SDFDiff: Differentiable Rendering of Signed
  Distance Fields for 3D Shape Optimization}. In \bibinfo{booktitle}{\emph{The
  IEEE/CVF Conference on Computer Vision and Pattern Recognition (CVPR)}}.
\newblock


\bibitem[\protect\citeauthoryear{Jin, Zhu, Geng, and Fedkiw}{Jin
  et~al\mbox{.}}{2018}]%
        {jin18}
\bibfield{author}{\bibinfo{person}{Ning Jin}, \bibinfo{person}{Yilin Zhu},
  \bibinfo{person}{Zhenglin Geng}, {and} \bibinfo{person}{Ronald Fedkiw}.}
  \bibinfo{year}{2018}\natexlab{}.
\newblock \bibinfo{title}{A Pixel-Based Framework for Data-Driven Clothing}.
\newblock
\newblock
\showeprint[arxiv]{1812.01677}~[cs.CV]


\bibitem[\protect\citeauthoryear{Kato, Ushiku, and Harada}{Kato
  et~al\mbox{.}}{2018}]%
        {kato18}
\bibfield{author}{\bibinfo{person}{Hiroharu Kato}, \bibinfo{person}{Yoshitaka
  Ushiku}, {and} \bibinfo{person}{Tatsuya Harada}.}
  \bibinfo{year}{2018}\natexlab{}.
\newblock \showarticletitle{Neural 3D Mesh Renderer}. In
  \bibinfo{booktitle}{\emph{The IEEE Conference on Computer Vision and Pattern
  Recognition (CVPR)}}.
\newblock


\bibitem[\protect\citeauthoryear{Kavan, Collins, {\v{Z}}{\'a}ra, and
  O'Sullivan}{Kavan et~al\mbox{.}}{2007}]%
        {kavan07}
\bibfield{author}{\bibinfo{person}{Ladislav Kavan}, \bibinfo{person}{Steven
  Collins}, \bibinfo{person}{Ji{\v{r}}{\'\i} {\v{Z}}{\'a}ra}, {and}
  \bibinfo{person}{Carol O'Sullivan}.} \bibinfo{year}{2007}\natexlab{}.
\newblock \showarticletitle{Skinning with dual quaternions}. In
  \bibinfo{booktitle}{\emph{Proceedings of the 2007 symposium on Interactive 3D
  graphics and games}}. ACM, \bibinfo{pages}{39--46}.
\newblock


\bibitem[\protect\citeauthoryear{Kim, Garrido, Tewari, Xu, Thies, Nie{\ss}ner,
  P{\'e}rez, Richardt, Zoll{\"o}fer, and Theobalt}{Kim et~al\mbox{.}}{2018a}]%
        {kim2018deep}
\bibfield{author}{\bibinfo{person}{Hyeongwoo Kim}, \bibinfo{person}{Pablo
  Garrido}, \bibinfo{person}{Ayush Tewari}, \bibinfo{person}{Weipeng Xu},
  \bibinfo{person}{Justus Thies}, \bibinfo{person}{Matthias Nie{\ss}ner},
  \bibinfo{person}{Patrick P{\'e}rez}, \bibinfo{person}{Christian Richardt},
  \bibinfo{person}{Michael Zoll{\"o}fer}, {and} \bibinfo{person}{Christian
  Theobalt}.} \bibinfo{year}{2018}\natexlab{a}.
\newblock \showarticletitle{Deep Video Portraits}.
\newblock \bibinfo{journal}{\emph{ACM Transactions on Graphics (TOG)}}
  \bibinfo{volume}{37}, \bibinfo{number}{4} (\bibinfo{year}{2018}),
  \bibinfo{pages}{163}.
\newblock


\bibitem[\protect\citeauthoryear{Kim, Garrido, Tewari, Xu, Thies, Nie{\ss}ner,
  P{\'e}rez, Richardt, Zoll{\"o}fer, and Theobalt}{Kim et~al\mbox{.}}{2018b}]%
        {kim18}
\bibfield{author}{\bibinfo{person}{Hyeongwoo Kim}, \bibinfo{person}{Pablo
  Garrido}, \bibinfo{person}{Ayush Tewari}, \bibinfo{person}{Weipeng Xu},
  \bibinfo{person}{Justus Thies}, \bibinfo{person}{Matthias Nie{\ss}ner},
  \bibinfo{person}{Patrick P{\'e}rez}, \bibinfo{person}{Christian Richardt},
  \bibinfo{person}{Michael Zoll{\"o}fer}, {and} \bibinfo{person}{Christian
  Theobalt}.} \bibinfo{year}{2018}\natexlab{b}.
\newblock \showarticletitle{Deep Video Portraits}.
\newblock \bibinfo{journal}{\emph{ACM Transactions on Graphics (TOG)}}
  \bibinfo{volume}{37}, \bibinfo{number}{4} (\bibinfo{year}{2018}),
  \bibinfo{pages}{163}.
\newblock


\bibitem[\protect\citeauthoryear{Kim and Vendrovsky}{Kim and
  Vendrovsky}{2008}]%
        {kim08}
\bibfield{author}{\bibinfo{person}{Tae-Yong Kim} {and} \bibinfo{person}{Eugene
  Vendrovsky}.} \bibinfo{year}{2008}\natexlab{}.
\newblock \showarticletitle{DrivenShape: A Data-Driven Approach for Shape
  Deformation}. In \bibinfo{booktitle}{\emph{ACM SIGGRAPH 2008 Talks}} (Los
  Angeles, California) \emph{(\bibinfo{series}{SIGGRAPH '08})}.
  \bibinfo{publisher}{Association for Computing Machinery},
  \bibinfo{address}{New York, NY, USA}, Article \bibinfo{articleno}{69},
  \bibinfo{numpages}{1}~pages.
\newblock
\showISBNx{9781605583433}
\urldef\tempurl%
\url{https://doi.org/10.1145/1401032.1401121}
\showDOI{\tempurl}


\bibitem[\protect\citeauthoryear{Kingma and Ba}{Kingma and Ba}{2014}]%
        {kingma14}
\bibfield{author}{\bibinfo{person}{Diederik Kingma} {and}
  \bibinfo{person}{Jimmy Ba}.} \bibinfo{year}{2014}\natexlab{}.
\newblock \showarticletitle{Adam: A Method for Stochastic Optimization}.
\newblock \bibinfo{journal}{\emph{International Conference on Learning
  Representations}} (\bibinfo{date}{12} \bibinfo{year}{2014}).
\newblock


\bibitem[\protect\citeauthoryear{L{\"a}hner, Cremers, and Tung}{L{\"a}hner
  et~al\mbox{.}}{2018}]%
        {lhner18}
\bibfield{author}{\bibinfo{person}{Zorah L{\"a}hner}, \bibinfo{person}{Daniel
  Cremers}, {and} \bibinfo{person}{Tony Tung}.}
  \bibinfo{year}{2018}\natexlab{}.
\newblock \showarticletitle{DeepWrinkles: Accurate and Realistic Clothing
  Modeling}.
\newblock \bibinfo{journal}{\emph{ArXiv}}  \bibinfo{volume}{abs/1808.03417}
  (\bibinfo{year}{2018}).
\newblock


\bibitem[\protect\citeauthoryear{Lambert}{Lambert}{1760}]%
        {lambert1760}
\bibfield{author}{\bibinfo{person}{J.~H. Lambert}.}
  \bibinfo{year}{1760}\natexlab{}.
\newblock \showarticletitle{{Photometria sive de mensure de gratibus luminis,
  colorum umbrae}}. In \bibinfo{booktitle}{\emph{Photometria sive de mensure de
  gratibus luminis, colorum umbrae, Eberhard Klett}}.
\newblock


\bibitem[\protect\citeauthoryear{Li, Liu, and Dai}{Li et~al\mbox{.}}{2014}]%
        {Li:2014}
\bibfield{author}{\bibinfo{person}{Guannan Li}, \bibinfo{person}{Yebin Liu},
  {and} \bibinfo{person}{Qionghai Dai}.} \bibinfo{year}{2014}\natexlab{}.
\newblock \showarticletitle{Free-viewpoint Video Relighting from Multi-view
  Sequence Under General Illumination}.
\newblock \bibinfo{journal}{\emph{Mach. Vision Appl.}} \bibinfo{volume}{25},
  \bibinfo{number}{7} (\bibinfo{date}{Oct.} \bibinfo{year}{2014}),
  \bibinfo{pages}{1737--1746}.
\newblock
\showISSN{0932-8092}
\urldef\tempurl%
\url{https://doi.org/10.1007/s00138-013-0559-0}
\showDOI{\tempurl}


\bibitem[\protect\citeauthoryear{Li, Niklaus, Snavely, and Wang}{Li
  et~al\mbox{.}}{2020}]%
        {Li20arxiv_nsff}
\bibfield{author}{\bibinfo{person}{Zhengqi Li}, \bibinfo{person}{Simon
  Niklaus}, \bibinfo{person}{Noah Snavely}, {and} \bibinfo{person}{Oliver
  Wang}.} \bibinfo{year}{2020}\natexlab{}.
\newblock \showarticletitle{Neural Scene Flow Fields for Space-Time View
  Synthesis of Dynamic Scenes}.
\newblock \bibinfo{journal}{\emph{https://arxiv.org/abs/2011.13084}}
  (\bibinfo{year}{2020}).
\newblock


\bibitem[\protect\citeauthoryear{Liang, Lin, and Koltun}{Liang
  et~al\mbox{.}}{2019}]%
        {liang19}
\bibfield{author}{\bibinfo{person}{Junbang Liang}, \bibinfo{person}{Ming~C.
  Lin}, {and} \bibinfo{person}{Vladlen Koltun}.}
  \bibinfo{year}{2019}\natexlab{}.
\newblock \showarticletitle{Differentiable Cloth Simulation for Inverse
  Problems}. In \bibinfo{booktitle}{\emph{Conference on Neural Information
  Processing Systems (NeurIPS)}}.
\newblock


\bibitem[\protect\citeauthoryear{Liu, Gu, Lin, Chua, and Theobalt}{Liu
  et~al\mbox{.}}{2020a}]%
        {liu2020neural}
\bibfield{author}{\bibinfo{person}{Lingjie Liu}, \bibinfo{person}{Jiatao Gu},
  \bibinfo{person}{Kyaw~Zaw Lin}, \bibinfo{person}{Tat-Seng Chua}, {and}
  \bibinfo{person}{Christian Theobalt}.} \bibinfo{year}{2020}\natexlab{a}.
\newblock \showarticletitle{Neural Sparse Voxel Fields}.
\newblock \bibinfo{journal}{\emph{NeurIPS}} (\bibinfo{year}{2020}).
\newblock


\bibitem[\protect\citeauthoryear{Liu, Xu, Habermann, Zollhoefer, Bernard, Kim,
  Wang, and Theobalt}{Liu et~al\mbox{.}}{2020b}]%
        {liu20}
\bibfield{author}{\bibinfo{person}{Lingjie Liu}, \bibinfo{person}{Weipeng Xu},
  \bibinfo{person}{Marc Habermann}, \bibinfo{person}{Michael Zollhoefer},
  \bibinfo{person}{Florian Bernard}, \bibinfo{person}{Hyeongwoo Kim},
  \bibinfo{person}{Wenping Wang}, {and} \bibinfo{person}{Christian Theobalt}.}
  \bibinfo{year}{2020}\natexlab{b}.
\newblock \bibinfo{title}{Neural Human Video Rendering by Learning Dynamic
  Textures and Rendering-to-Video Translation}.
\newblock
\newblock
\showeprint[arxiv]{2001.04947}~[cs.GR]


\bibitem[\protect\citeauthoryear{Liu, Xu, Zollh\"{o}fer, Kim, Bernard,
  Habermann, Wang, and Theobalt}{Liu et~al\mbox{.}}{2019b}]%
        {liu19}
\bibfield{author}{\bibinfo{person}{Lingjie Liu}, \bibinfo{person}{Weipeng Xu},
  \bibinfo{person}{Michael Zollh\"{o}fer}, \bibinfo{person}{Hyeongwoo Kim},
  \bibinfo{person}{Florian Bernard}, \bibinfo{person}{Marc Habermann},
  \bibinfo{person}{Wenping Wang}, {and} \bibinfo{person}{Christian Theobalt}.}
  \bibinfo{year}{2019}\natexlab{b}.
\newblock \showarticletitle{Neural Rendering and Reenactment of Human Actor
  Videos}.
\newblock \bibinfo{journal}{\emph{ACM Trans. Graph.}} \bibinfo{volume}{38},
  \bibinfo{number}{5}, Article \bibinfo{articleno}{139} (\bibinfo{date}{Oct.}
  \bibinfo{year}{2019}), \bibinfo{numpages}{14}~pages.
\newblock
\showISSN{0730-0301}
\urldef\tempurl%
\url{https://doi.org/10.1145/3333002}
\showDOI{\tempurl}


\bibitem[\protect\citeauthoryear{Liu, Li, Chen, and Li}{Liu
  et~al\mbox{.}}{2019a}]%
        {liu19a}
\bibfield{author}{\bibinfo{person}{Shichen Liu}, \bibinfo{person}{Tianye Li},
  \bibinfo{person}{Weikai Chen}, {and} \bibinfo{person}{Hao Li}.}
  \bibinfo{year}{2019}\natexlab{a}.
\newblock \showarticletitle{Soft Rasterizer: A Differentiable Renderer for
  Image-based 3D Reasoning}.
\newblock \bibinfo{journal}{\emph{The IEEE International Conference on Computer
  Vision (ICCV)}} (\bibinfo{date}{Oct} \bibinfo{year}{2019}).
\newblock


\bibitem[\protect\citeauthoryear{Lombardi, Saragih, Simon, and Sheikh}{Lombardi
  et~al\mbox{.}}{2018}]%
        {lombardi18}
\bibfield{author}{\bibinfo{person}{Stephen Lombardi}, \bibinfo{person}{Jason
  Saragih}, \bibinfo{person}{Tomas Simon}, {and} \bibinfo{person}{Yaser
  Sheikh}.} \bibinfo{year}{2018}\natexlab{}.
\newblock \showarticletitle{Deep appearance models for face rendering}.
\newblock \bibinfo{journal}{\emph{ACM Transactions on Graphics}}
  \bibinfo{volume}{37}, \bibinfo{number}{4} (\bibinfo{date}{Aug}
  \bibinfo{year}{2018}), \bibinfo{pages}{1–13}.
\newblock
\showISSN{1557-7368}
\urldef\tempurl%
\url{https://doi.org/10.1145/3197517.3201401}
\showDOI{\tempurl}


\bibitem[\protect\citeauthoryear{Lombardi, Simon, Saragih, Schwartz, Lehrmann,
  and Sheikh}{Lombardi et~al\mbox{.}}{2019}]%
        {lombardi2019neural}
\bibfield{author}{\bibinfo{person}{Stephen Lombardi}, \bibinfo{person}{Tomas
  Simon}, \bibinfo{person}{Jason Saragih}, \bibinfo{person}{Gabriel Schwartz},
  \bibinfo{person}{Andreas Lehrmann}, {and} \bibinfo{person}{Yaser Sheikh}.}
  \bibinfo{year}{2019}\natexlab{}.
\newblock \showarticletitle{Neural volumes: Learning dynamic renderable volumes
  from images}.
\newblock \bibinfo{journal}{\emph{ACM Transactions on Graphics (TOG)}}
  \bibinfo{volume}{38}, \bibinfo{number}{4} (\bibinfo{year}{2019}),
  \bibinfo{pages}{65}.
\newblock


\bibitem[\protect\citeauthoryear{Loper, Mahmood, Romero, Pons-Moll, and
  Black}{Loper et~al\mbox{.}}{2015}]%
        {loper15}
\bibfield{author}{\bibinfo{person}{Matthew Loper}, \bibinfo{person}{Naureen
  Mahmood}, \bibinfo{person}{Javier Romero}, \bibinfo{person}{Gerard
  Pons-Moll}, {and} \bibinfo{person}{Michael~J. Black}.}
  \bibinfo{year}{2015}\natexlab{}.
\newblock \showarticletitle{{SMPL}: A Skinned Multi-Person Linear Model}.
\newblock \bibinfo{journal}{\emph{ACM Trans. Graphics (Proc. SIGGRAPH Asia)}}
  \bibinfo{volume}{34}, \bibinfo{number}{6} (\bibinfo{date}{Oct.}
  \bibinfo{year}{2015}), \bibinfo{pages}{248:1--248:16}.
\newblock


\bibitem[\protect\citeauthoryear{Loper and Black}{Loper and Black}{2014}]%
        {loper14a}
\bibfield{author}{\bibinfo{person}{Matthew~M. Loper} {and}
  \bibinfo{person}{Michael~J. Black}.} \bibinfo{year}{2014}\natexlab{}.
\newblock \showarticletitle{OpenDR: An Approximate Differentiable Renderer}. In
  \bibinfo{booktitle}{\emph{Computer Vision -- ECCV 2014}},
  \bibfield{editor}{\bibinfo{person}{David Fleet}, \bibinfo{person}{Tomas
  Pajdla}, \bibinfo{person}{Bernt Schiele}, {and} \bibinfo{person}{Tinne
  Tuytelaars}} (Eds.). \bibinfo{publisher}{Springer International Publishing},
  \bibinfo{address}{Cham}, \bibinfo{pages}{154--169}.
\newblock


\bibitem[\protect\citeauthoryear{Ma, Jia, Sun, Schiele, Tuytelaars, and
  Van~Gool}{Ma et~al\mbox{.}}{2017}]%
        {MaSJSTV2017}
\bibfield{author}{\bibinfo{person}{Liqian Ma}, \bibinfo{person}{Xu Jia},
  \bibinfo{person}{Qianru Sun}, \bibinfo{person}{Bernt Schiele},
  \bibinfo{person}{Tinne Tuytelaars}, {and} \bibinfo{person}{Luc Van~Gool}.}
  \bibinfo{year}{2017}\natexlab{}.
\newblock \showarticletitle{Pose guided person image generation}. In
  \bibinfo{booktitle}{\emph{Advances in Neural Information Processing
  Systems}}. \bibinfo{pages}{405--415}.
\newblock


\bibitem[\protect\citeauthoryear{Ma, Sun, Georgoulis, van Gool, Schiele, and
  Fritz}{Ma et~al\mbox{.}}{2018}]%
        {Ma18}
\bibfield{author}{\bibinfo{person}{Liqian Ma}, \bibinfo{person}{Qianru Sun},
  \bibinfo{person}{Stamatios Georgoulis}, \bibinfo{person}{Luc van Gool},
  \bibinfo{person}{Bernt Schiele}, {and} \bibinfo{person}{Mario Fritz}.}
  \bibinfo{year}{2018}\natexlab{}.
\newblock \showarticletitle{Disentangled Person Image Generation}.
\newblock \bibinfo{journal}{\emph{Computer Vision and Pattern Recognition
  (CVPR)}} (\bibinfo{year}{2018}).
\newblock


\bibitem[\protect\citeauthoryear{Ma, Yang, Ranjan, Pujades, Pons-Moll, Tang,
  and Black}{Ma et~al\mbox{.}}{2020}]%
        {ma20autoenclother}
\bibfield{author}{\bibinfo{person}{Qianli Ma}, \bibinfo{person}{Jinlong Yang},
  \bibinfo{person}{Anurag Ranjan}, \bibinfo{person}{Sergi Pujades},
  \bibinfo{person}{Gerard Pons-Moll}, \bibinfo{person}{Siyu Tang}, {and}
  \bibinfo{person}{Michael Black}.} \bibinfo{year}{2020}\natexlab{}.
\newblock \showarticletitle{Learning to Dress 3D People in Generative
  Clothing}. In \bibinfo{booktitle}{\emph{{IEEE} Conference on Computer Vision
  and Pattern Recognition (CVPR)}}. {IEEE}.
\newblock


\bibitem[\protect\citeauthoryear{Magnenat-Thalmann, Laperri{`{e}}re, and
  Thalmann}{Magnenat-Thalmann et~al\mbox{.}}{1988}]%
        {Magnenat-Thalmann1988:4}
\bibfield{author}{\bibinfo{person}{N. Magnenat-Thalmann}, \bibinfo{person}{A.
  Laperri{`{e}}re}, {and} \bibinfo{person}{D. Thalmann}.}
  \bibinfo{year}{1988}\natexlab{}.
\newblock \showarticletitle{Joint-Dependent Local Deformations for Hand
  Animation and Object Grasping}. In \bibinfo{booktitle}{\emph{Proceedings of
  Graphics Interface '88}} (Edmonton, Alberta, Canada)
  \emph{(\bibinfo{series}{GI '88})}. \bibinfo{publisher}{Canadian Man-Computer
  Communications Society}, \bibinfo{address}{Toronto, Ontario, Canada},
  \bibinfo{pages}{26--33}.
\newblock
\showISSN{0713-5424}
\urldef\tempurl%
\url{http://graphicsinterface.org/wp-content/uploads/gi1988-4.pdf}
\showURL{%
\tempurl}


\bibitem[\protect\citeauthoryear{Martin-Brualla, Pandey, Yang, Pidlypenskyi,
  Taylor, Valentin, Khamis, Davidson, Tkach, Lincoln, Kowdle, Rhemann, Goldman,
  Keskin, Seitz, Izadi, and Fanello}{Martin-Brualla et~al\mbox{.}}{2018}]%
        {Martin18}
\bibfield{author}{\bibinfo{person}{Ricardo Martin-Brualla},
  \bibinfo{person}{Rohit Pandey}, \bibinfo{person}{Shuoran Yang},
  \bibinfo{person}{Pavel Pidlypenskyi}, \bibinfo{person}{Jonathan Taylor},
  \bibinfo{person}{Julien Valentin}, \bibinfo{person}{Sameh Khamis},
  \bibinfo{person}{Philip Davidson}, \bibinfo{person}{Anastasia Tkach},
  \bibinfo{person}{Peter Lincoln}, \bibinfo{person}{Adarsh Kowdle},
  \bibinfo{person}{Christoph Rhemann}, \bibinfo{person}{Dan~B Goldman},
  \bibinfo{person}{Cem Keskin}, \bibinfo{person}{Steve Seitz},
  \bibinfo{person}{Shahram Izadi}, {and} \bibinfo{person}{Sean Fanello}.}
  \bibinfo{year}{2018}\natexlab{}.
\newblock \showarticletitle{<i>LookinGood</i>: Enhancing Performance Capture
  with Real-Time Neural Re-Rendering}.
\newblock \bibinfo{journal}{\emph{ACM Trans. Graph.}} \bibinfo{volume}{37},
  \bibinfo{number}{6}, Article \bibinfo{articleno}{255} (\bibinfo{date}{Dec.}
  \bibinfo{year}{2018}), \bibinfo{numpages}{14}~pages.
\newblock
\showISSN{0730-0301}
\urldef\tempurl%
\url{https://doi.org/10.1145/3272127.3275099}
\showDOI{\tempurl}


\bibitem[\protect\citeauthoryear{Mildenhall, Srinivasan, Tancik, Barron,
  Ramamoorthi, and Ng}{Mildenhall et~al\mbox{.}}{2020}]%
        {mildenhall2020nerf}
\bibfield{author}{\bibinfo{person}{Ben Mildenhall}, \bibinfo{person}{Pratul~P.
  Srinivasan}, \bibinfo{person}{Matthew Tancik}, \bibinfo{person}{Jonathan~T.
  Barron}, \bibinfo{person}{Ravi Ramamoorthi}, {and} \bibinfo{person}{Ren Ng}.}
  \bibinfo{year}{2020}\natexlab{}.
\newblock \showarticletitle{NeRF: Representing Scenes as Neural Radiance Fields
  for View Synthesis}. In \bibinfo{booktitle}{\emph{ECCV}}.
\newblock


\bibitem[\protect\citeauthoryear{Mueller}{Mueller}{1966}]%
        {mueller66}
\bibfield{author}{\bibinfo{person}{Claus Mueller}.}
  \bibinfo{year}{1966}\natexlab{}.
\newblock \showarticletitle{{Spherical Harmonics}}. In
  \bibinfo{booktitle}{\emph{Spherical Harmonics, Springer}}.
\newblock


\bibitem[\protect\citeauthoryear{Narain, Samii, and O'Brien}{Narain
  et~al\mbox{.}}{2012}]%
        {Narain12}
\bibfield{author}{\bibinfo{person}{Rahul Narain}, \bibinfo{person}{Armin
  Samii}, {and} \bibinfo{person}{James~F. O'Brien}.}
  \bibinfo{year}{2012}\natexlab{}.
\newblock \showarticletitle{Adaptive Anisotropic Remeshing for Cloth
  Simulation}.
\newblock \bibinfo{journal}{\emph{ACM Transactions on Graphics}}
  \bibinfo{volume}{31}, \bibinfo{number}{6} (\bibinfo{date}{Nov.}
  \bibinfo{year}{2012}), \bibinfo{pages}{147:1--10}.
\newblock
\urldef\tempurl%
\url{http://graphics.berkeley.edu/papers/Narain-AAR-2012-11/}
\showURL{%
\tempurl}
\newblock
\shownote{Proceedings of ACM SIGGRAPH Asia 2012, Singapore.}


\bibitem[\protect\citeauthoryear{Nealen, Müller, Keiser, Boxerman, and
  Carlson}{Nealen et~al\mbox{.}}{2005}]%
        {Nealen05}
\bibfield{author}{\bibinfo{person}{A. Nealen}, \bibinfo{person}{Matthias
  Müller}, \bibinfo{person}{Richard Keiser}, \bibinfo{person}{Eddy Boxerman},
  {and} \bibinfo{person}{M. Carlson}.} \bibinfo{year}{2005}\natexlab{}.
\newblock \showarticletitle{Physically based deformable models in computer
  graphics}.
\newblock \bibinfo{journal}{\emph{Eurographics: State of the Art Report}}
  (\bibinfo{date}{01} \bibinfo{year}{2005}), \bibinfo{pages}{71--94}.
\newblock


\bibitem[\protect\citeauthoryear{Park, Sinha, Barron, Bouaziz, Goldman, Seitz,
  and Martin-Brualla}{Park et~al\mbox{.}}{2020}]%
        {Park20arxiv_nerfies}
\bibfield{author}{\bibinfo{person}{Keunhong Park}, \bibinfo{person}{Utkarsh
  Sinha}, \bibinfo{person}{Jonathan Barron}, \bibinfo{person}{Sofien Bouaziz},
  \bibinfo{person}{Dan Goldman}, \bibinfo{person}{Steven Seitz}, {and}
  \bibinfo{person}{Ricardo Martin-Brualla}.} \bibinfo{year}{2020}\natexlab{}.
\newblock \showarticletitle{Deformable Neural Radiance Fields}.
\newblock \bibinfo{journal}{\emph{https://arxiv.org/abs/2011.12948}}
  (\bibinfo{year}{2020}).
\newblock


\bibitem[\protect\citeauthoryear{Patel, Liao, and Pons-Moll}{Patel
  et~al\mbox{.}}{2020}]%
        {patel20}
\bibfield{author}{\bibinfo{person}{Chaitanya Patel},
  \bibinfo{person}{Zhouyingcheng Liao}, {and} \bibinfo{person}{Gerard
  Pons-Moll}.} \bibinfo{year}{2020}\natexlab{}.
\newblock \showarticletitle{TailorNet: Predicting Clothing in 3D as a Function
  of Human Pose, Shape and Garment Style}. In \bibinfo{booktitle}{\emph{{IEEE}
  Conference on Computer Vision and Pattern Recognition (CVPR)}}. {IEEE}.
\newblock


\bibitem[\protect\citeauthoryear{{{Photo Scan}}}{{{Photo Scan}}}{2016}]%
        {agisoftphotoscan}
{{Photo Scan}} \bibinfo{year}{2016}\natexlab{}.
\newblock \bibinfo{title}{{PhotoScan}}.
\newblock \bibinfo{howpublished}{\url{http://www.agisoft.com}}.
\newblock


\bibitem[\protect\citeauthoryear{Pineda}{Pineda}{1988}]%
        {pineda88}
\bibfield{author}{\bibinfo{person}{Juan Pineda}.}
  \bibinfo{year}{1988}\natexlab{}.
\newblock \showarticletitle{A Parallel Algorithm for Polygon Rasterization}.
\newblock \bibinfo{journal}{\emph{SIGGRAPH Comput. Graph.}}
  \bibinfo{volume}{22}, \bibinfo{number}{4} (\bibinfo{date}{June}
  \bibinfo{year}{1988}), \bibinfo{pages}{17–20}.
\newblock
\showISSN{0097-8930}
\urldef\tempurl%
\url{https://doi.org/10.1145/378456.378457}
\showDOI{\tempurl}


\bibitem[\protect\citeauthoryear{Pons-Moll, Pujades, Hu, and Black}{Pons-Moll
  et~al\mbox{.}}{2017}]%
        {pons17}
\bibfield{author}{\bibinfo{person}{Gerard Pons-Moll}, \bibinfo{person}{Sergi
  Pujades}, \bibinfo{person}{Sonny Hu}, {and} \bibinfo{person}{Michael Black}.}
  \bibinfo{year}{2017}\natexlab{}.
\newblock \showarticletitle{{ClothCap}: Seamless {4D} Clothing Capture and
  Retargeting}.
\newblock \bibinfo{journal}{\emph{ACM Transactions on Graphics, (Proc.
  SIGGRAPH)}} \bibinfo{volume}{36}, \bibinfo{number}{4} (\bibinfo{year}{2017}).
\newblock
\urldef\tempurl%
\url{http://dx.doi.org/10.1145/3072959.3073711}
\showURL{%
\tempurl}


\bibitem[\protect\citeauthoryear{Pumarola, Agudo, Sanfeliu, and
  Moreno-Noguer}{Pumarola et~al\mbox{.}}{2018}]%
        {Pumarola_2018_CVPR}
\bibfield{author}{\bibinfo{person}{Albert Pumarola}, \bibinfo{person}{Antonio
  Agudo}, \bibinfo{person}{Alberto Sanfeliu}, {and} \bibinfo{person}{Francesc
  Moreno-Noguer}.} \bibinfo{year}{2018}\natexlab{}.
\newblock \showarticletitle{Unsupervised Person Image Synthesis in Arbitrary
  Poses}. In \bibinfo{booktitle}{\emph{The IEEE Conference on Computer Vision
  and Pattern Recognition (CVPR)}}.
\newblock


\bibitem[\protect\citeauthoryear{Pumarola, Corona, Pons-Moll, and
  Moreno-Noguer}{Pumarola et~al\mbox{.}}{2020}]%
        {Pumarola20arxiv_D_NeRF}
\bibfield{author}{\bibinfo{person}{Albert Pumarola}, \bibinfo{person}{Enric
  Corona}, \bibinfo{person}{Gerard Pons-Moll}, {and} \bibinfo{person}{Francesc
  Moreno-Noguer}.} \bibinfo{year}{2020}\natexlab{}.
\newblock \showarticletitle{{D-NeRF}: Neural Radiance Fields for Dynamic
  Scenes}.
\newblock \bibinfo{journal}{\emph{https://arxiv.org/abs/2011.13961}}
  (\bibinfo{year}{2020}).
\newblock


\bibitem[\protect\citeauthoryear{Raj, Zollhoefer, Simon, Saragih, Saito, Hays,
  and Lombardi}{Raj et~al\mbox{.}}{2020}]%
        {raj2020pva}
\bibfield{author}{\bibinfo{person}{Amit Raj}, \bibinfo{person}{Michael
  Zollhoefer}, \bibinfo{person}{Tomas Simon}, \bibinfo{person}{Jason Saragih},
  \bibinfo{person}{Shunsuke Saito}, \bibinfo{person}{James Hays}, {and}
  \bibinfo{person}{Stephen Lombardi}.} \bibinfo{year}{2020}\natexlab{}.
\newblock \showarticletitle{PVA: Pixel-aligned Volumetric Avatars}. In
  \bibinfo{booktitle}{\emph{arXiv:2101.02697}}.
\newblock


\bibitem[\protect\citeauthoryear{Santesteban, Otaduy, and Casas}{Santesteban
  et~al\mbox{.}}{2019}]%
        {santesteban19}
\bibfield{author}{\bibinfo{person}{Igor Santesteban},
  \bibinfo{person}{Miguel~A. Otaduy}, {and} \bibinfo{person}{Dan Casas}.}
  \bibinfo{year}{2019}\natexlab{}.
\newblock \showarticletitle{Learning-Based Animation of Clothing for Virtual
  Try-On}.
\newblock \bibinfo{journal}{\emph{Comput. Graph. Forum}}  \bibinfo{volume}{38}
  (\bibinfo{year}{2019}), \bibinfo{pages}{355--366}.
\newblock


\bibitem[\protect\citeauthoryear{Sarkar, Mehta, Xu, Golyanik, and
  Theobalt}{Sarkar et~al\mbox{.}}{2020}]%
        {Sarkar2020}
\bibfield{author}{\bibinfo{person}{Kripasindhu Sarkar},
  \bibinfo{person}{Dushyant Mehta}, \bibinfo{person}{Weipeng Xu},
  \bibinfo{person}{Vladislav Golyanik}, {and} \bibinfo{person}{Christian
  Theobalt}.} \bibinfo{year}{2020}\natexlab{}.
\newblock \showarticletitle{Neural Re-Rendering of Humans from a Single Image}.
  In \bibinfo{booktitle}{\emph{European Conference on Computer Vision (ECCV)}}.
\newblock


\bibitem[\protect\citeauthoryear{Shysheya, Zakharov, Aliev, Bashirov, Burkov,
  Iskakov, Ivakhnenko, Malkov, Pasechnik, Ulyanov, Vakhitov, and
  Lempitsky}{Shysheya et~al\mbox{.}}{2019}]%
        {shysheya19}
\bibfield{author}{\bibinfo{person}{Aliaksandra Shysheya}, \bibinfo{person}{Egor
  Zakharov}, \bibinfo{person}{Kara-Ali Aliev}, \bibinfo{person}{Renat
  Bashirov}, \bibinfo{person}{Egor Burkov}, \bibinfo{person}{Karim Iskakov},
  \bibinfo{person}{Aleksei Ivakhnenko}, \bibinfo{person}{Yury Malkov},
  \bibinfo{person}{Igor Pasechnik}, \bibinfo{person}{Dmitry Ulyanov},
  \bibinfo{person}{Alexander Vakhitov}, {and} \bibinfo{person}{Victor
  Lempitsky}.} \bibinfo{year}{2019}\natexlab{}.
\newblock \bibinfo{title}{Textured Neural Avatars}.
\newblock
\newblock
\showeprint[arxiv]{1905.08776}~[cs.CV]


\bibitem[\protect\citeauthoryear{Si, Wang, Wang, and Tan}{Si
  et~al\mbox{.}}{2018}]%
        {Si_2018_CVPR}
\bibfield{author}{\bibinfo{person}{Chenyang Si}, \bibinfo{person}{Wei Wang},
  \bibinfo{person}{Liang Wang}, {and} \bibinfo{person}{Tieniu Tan}.}
  \bibinfo{year}{2018}\natexlab{}.
\newblock \showarticletitle{Multistage Adversarial Losses for Pose-Based Human
  Image Synthesis}. In \bibinfo{booktitle}{\emph{The IEEE Conference on
  Computer Vision and Pattern Recognition (CVPR)}}.
\newblock


\bibitem[\protect\citeauthoryear{Siarohin, Sangineto, Lathuiliere, and
  Sebe}{Siarohin et~al\mbox{.}}{2018}]%
        {SiaroSLS2017}
\bibfield{author}{\bibinfo{person}{Aliaksandr Siarohin}, \bibinfo{person}{Enver
  Sangineto}, \bibinfo{person}{Stephane Lathuiliere}, {and}
  \bibinfo{person}{Nicu Sebe}.} \bibinfo{year}{2018}\natexlab{}.
\newblock \showarticletitle{Deformable {GANs} for Pose-based Human Image
  Generation}. In \bibinfo{booktitle}{\emph{CVPR 2018}}.
\newblock


\bibitem[\protect\citeauthoryear{Sida~Peng}{Sida~Peng}{2020}]%
        {peng2020neural}
\bibfield{author}{\bibinfo{person}{Yinghao Xu Qianqian Wang Qing Shuai Hujun
  Bao Xiaowei~Zhou Sida~Peng, Yuanqing~Zhang}.}
  \bibinfo{year}{2020}\natexlab{}.
\newblock \showarticletitle{Neural Body: Implicit Neural Representations with
  Structured Latent Codes for Novel View Synthesis of Dynamic Humans}.
\newblock \bibinfo{journal}{\emph{arXiv preprint arXiv:2012.15838}}
  (\bibinfo{year}{2020}).
\newblock


\bibitem[\protect\citeauthoryear{Sitzmann, Thies, Heide, Nie{\ss}ner,
  Wetzstein, and Zollh{\"o}fer}{Sitzmann et~al\mbox{.}}{2019a}]%
        {sitzmann2019deepvoxels}
\bibfield{author}{\bibinfo{person}{Vincent Sitzmann}, \bibinfo{person}{Justus
  Thies}, \bibinfo{person}{Felix Heide}, \bibinfo{person}{Matthias
  Nie{\ss}ner}, \bibinfo{person}{Gordon Wetzstein}, {and}
  \bibinfo{person}{Michael Zollh{\"o}fer}.} \bibinfo{year}{2019}\natexlab{a}.
\newblock \showarticletitle{DeepVoxels: Learning Persistent 3D Feature
  Embeddings}. In \bibinfo{booktitle}{\emph{Computer Vision and Pattern
  Recognition (CVPR)}}.
\newblock


\bibitem[\protect\citeauthoryear{Sitzmann, Zollh{\"o}fer, and
  Wetzstein}{Sitzmann et~al\mbox{.}}{2019b}]%
        {sitzmann2019srns}
\bibfield{author}{\bibinfo{person}{Vincent Sitzmann}, \bibinfo{person}{Michael
  Zollh{\"o}fer}, {and} \bibinfo{person}{Gordon Wetzstein}.}
  \bibinfo{year}{2019}\natexlab{b}.
\newblock \showarticletitle{Scene Representation Networks: Continuous
  3D-Structure-Aware Neural Scene Representations}. In
  \bibinfo{booktitle}{\emph{Advances in Neural Information Processing
  Systems}}.
\newblock


\bibitem[\protect\citeauthoryear{Sorkine and Alexa}{Sorkine and Alexa}{2007}]%
        {sorkine07}
\bibfield{author}{\bibinfo{person}{Olga Sorkine} {and} \bibinfo{person}{Marc
  Alexa}.} \bibinfo{year}{2007}\natexlab{}.
\newblock \showarticletitle{As-rigid-as-possible Surface Modeling}. In
  \bibinfo{booktitle}{\emph{Proceedings of the Fifth Eurographics Symposium on
  Geometry Processing}} (Barcelona, Spain) \emph{(\bibinfo{series}{SGP '07})}.
  \bibinfo{publisher}{Eurographics Association}.
\newblock


\bibitem[\protect\citeauthoryear{Starke, Zhang, Komura, and Saito}{Starke
  et~al\mbox{.}}{2019}]%
        {Starke2019NeuralSM}
\bibfield{author}{\bibinfo{person}{S. Starke}, \bibinfo{person}{H. Zhang},
  \bibinfo{person}{T. Komura}, {and} \bibinfo{person}{J. Saito}.}
  \bibinfo{year}{2019}\natexlab{}.
\newblock \showarticletitle{Neural state machine for character-scene
  interactions}.
\newblock \bibinfo{journal}{\emph{ACM Transactions on Graphics (TOG)}}
  \bibinfo{volume}{38} (\bibinfo{year}{2019}), \bibinfo{pages}{1 -- 14}.
\newblock


\bibitem[\protect\citeauthoryear{Stoll, Gall, de~Aguiar, Thrun, and
  Theobalt}{Stoll et~al\mbox{.}}{2010}]%
        {stoll10}
\bibfield{author}{\bibinfo{person}{Carsten Stoll}, \bibinfo{person}{Juergen
  Gall}, \bibinfo{person}{Edilson de Aguiar}, \bibinfo{person}{Sebastian
  Thrun}, {and} \bibinfo{person}{Christian Theobalt}.}
  \bibinfo{year}{2010}\natexlab{}.
\newblock \showarticletitle{Video-Based Reconstruction of Animatable Human
  Characters}.
\newblock \bibinfo{journal}{\emph{ACM Trans. Graph.}} \bibinfo{volume}{29},
  \bibinfo{number}{6}, Article \bibinfo{articleno}{139} (\bibinfo{date}{Dec.}
  \bibinfo{year}{2010}), \bibinfo{numpages}{10}~pages.
\newblock
\showISSN{0730-0301}
\urldef\tempurl%
\url{https://doi.org/10.1145/1882261.1866161}
\showDOI{\tempurl}


\bibitem[\protect\citeauthoryear{Su, Wan, Yu, Liu, Fang, Wang, and Liu}{Su
  et~al\mbox{.}}{2020}]%
        {su20}
\bibfield{author}{\bibinfo{person}{Zhaoqi Su}, \bibinfo{person}{Weilin Wan},
  \bibinfo{person}{Tao Yu}, \bibinfo{person}{Lingjie Liu}, \bibinfo{person}{Lu
  Fang}, \bibinfo{person}{Wenping Wang}, {and} \bibinfo{person}{Yebin Liu}.}
  \bibinfo{year}{2020}\natexlab{}.
\newblock \showarticletitle{MulayCap: Multi-layer Human Performance Capture
  Using A Monocular Video Camera}.
\newblock \bibinfo{journal}{\emph{IEEE Transactions on Visualization and
  Computer Graphics}} (\bibinfo{year}{2020}), \bibinfo{pages}{1–1}.
\newblock
\showISSN{2160-9306}
\urldef\tempurl%
\url{https://doi.org/10.1109/tvcg.2020.3027763}
\showDOI{\tempurl}


\bibitem[\protect\citeauthoryear{Sumner, Schmid, and Pauly}{Sumner
  et~al\mbox{.}}{2007}]%
        {sumner07}
\bibfield{author}{\bibinfo{person}{Robert~W. Sumner}, \bibinfo{person}{Johannes
  Schmid}, {and} \bibinfo{person}{Mark Pauly}.}
  \bibinfo{year}{2007}\natexlab{}.
\newblock \showarticletitle{Embedded Deformation for Shape Manipulation}.
\newblock \bibinfo{journal}{\emph{ACM Trans. Graph.}} \bibinfo{volume}{26},
  \bibinfo{number}{3} (\bibinfo{date}{July} \bibinfo{year}{2007}).
\newblock


\bibitem[\protect\citeauthoryear{Tang, wang, Liu, Tong, and Manocha}{Tang
  et~al\mbox{.}}{2018}]%
        {Tang18}
\bibfield{author}{\bibinfo{person}{Min Tang}, \bibinfo{person}{tongtong wang},
  \bibinfo{person}{Zhongyuan Liu}, \bibinfo{person}{Ruofeng Tong}, {and}
  \bibinfo{person}{Dinesh Manocha}.} \bibinfo{year}{2018}\natexlab{}.
\newblock \showarticletitle{I-Cloth: Incremental Collision Handling for
  GPU-Based Interactive Cloth Simulation}.
\newblock \bibinfo{journal}{\emph{ACM Trans. Graph.}} \bibinfo{volume}{37},
  \bibinfo{number}{6}, Article \bibinfo{articleno}{204} (\bibinfo{date}{Dec.}
  \bibinfo{year}{2018}), \bibinfo{numpages}{10}~pages.
\newblock
\showISSN{0730-0301}
\urldef\tempurl%
\url{https://doi.org/10.1145/3272127.3275005}
\showDOI{\tempurl}


\bibitem[\protect\citeauthoryear{Tao, Zheng, Zhong, Zhao, Quionhai, Pons-Moll,
  and Liu}{Tao et~al\mbox{.}}{2019}]%
        {tao19}
\bibfield{author}{\bibinfo{person}{Yu Tao}, \bibinfo{person}{Zerong Zheng},
  \bibinfo{person}{Yuan Zhong}, \bibinfo{person}{Jianhui Zhao},
  \bibinfo{person}{Dai Quionhai}, \bibinfo{person}{Gerard Pons-Moll}, {and}
  \bibinfo{person}{Yebin Liu}.} \bibinfo{year}{2019}\natexlab{}.
\newblock \showarticletitle{SimulCap : Single-View Human Performance Capture
  with Cloth Simulation}. In \bibinfo{booktitle}{\emph{{IEEE} Conference on
  Computer Vision and Pattern Recognition (CVPR)}}.
\newblock


\bibitem[\protect\citeauthoryear{{{TheCaptury}}}{{{TheCaptury}}}{2020}]%
        {captury}
{{TheCaptury}} \bibinfo{year}{2020}\natexlab{}.
\newblock \bibinfo{title}{{The Captury}}.
\newblock \bibinfo{howpublished}{\url{http://www.thecaptury.com/}}.
\newblock


\bibitem[\protect\citeauthoryear{Thies, Zollh\"{o}fer, and Nießner}{Thies
  et~al\mbox{.}}{2019}]%
        {Thies2019}
\bibfield{author}{\bibinfo{person}{Justus Thies}, \bibinfo{person}{Michael
  Zollh\"{o}fer}, {and} \bibinfo{person}{Matthias Nießner}.}
  \bibinfo{year}{2019}\natexlab{}.
\newblock \showarticletitle{Deferred neural rendering: image synthesis using
  neural textures}.
\newblock \bibinfo{journal}{\emph{ACM Transactions on Graphics}}
  \bibinfo{volume}{38} (\bibinfo{year}{2019}).
\newblock


\bibitem[\protect\citeauthoryear{{{Treedys}}}{{{Treedys}}}{2020}]%
        {treedys}
{{Treedys}} \bibinfo{year}{2020}\natexlab{}.
\newblock \bibinfo{title}{{Treedys}}.
\newblock \bibinfo{howpublished}{\url{https://www.treedys.com/}}.
\newblock


\bibitem[\protect\citeauthoryear{Tretschk, Tewari, Golyanik, Zollh{\"o}fer,
  Lassner, and Theobalt}{Tretschk et~al\mbox{.}}{2020}]%
        {Tretschk20arxiv_NR-NeRF}
\bibfield{author}{\bibinfo{person}{Edgar Tretschk}, \bibinfo{person}{Ayush
  Tewari}, \bibinfo{person}{Vladislav Golyanik}, \bibinfo{person}{Michael
  Zollh{\"o}fer}, \bibinfo{person}{Christoph Lassner}, {and}
  \bibinfo{person}{Christian Theobalt}.} \bibinfo{year}{2020}\natexlab{}.
\newblock \showarticletitle{Non-Rigid Neural Radiance Fields: Reconstruction
  and Novel View Synthesis of a Deforming Scene from Monocular Video}.
\newblock \bibinfo{journal}{\emph{https://arxiv.org/abs/2012.12247}}
  (\bibinfo{year}{2020}).
\newblock


\bibitem[\protect\citeauthoryear{Volino, Casas, Collomosse, and Hilton}{Volino
  et~al\mbox{.}}{2014}]%
        {Volino2014}
\bibfield{author}{\bibinfo{person}{Marco Volino}, \bibinfo{person}{Dan Casas},
  \bibinfo{person}{John Collomosse}, {and} \bibinfo{person}{Adrian Hilton}.}
  \bibinfo{year}{2014}\natexlab{}.
\newblock \showarticletitle{Optimal Representation of Multiple View Video}. In
  \bibinfo{booktitle}{\emph{Proceedings of the British Machine Vision
  Conference}}. \bibinfo{publisher}{BMVA Press}.
\newblock


\bibitem[\protect\citeauthoryear{Wang, Hecht, Ramamoorthi, and O'Brien}{Wang
  et~al\mbox{.}}{2010}]%
        {wang10}
\bibfield{author}{\bibinfo{person}{Huamin Wang}, \bibinfo{person}{Florian
  Hecht}, \bibinfo{person}{Ravi Ramamoorthi}, {and} \bibinfo{person}{James~F.
  O'Brien}.} \bibinfo{year}{2010}\natexlab{}.
\newblock \showarticletitle{Example-Based Wrinkle Synthesis for Clothing
  Animation}.
\newblock \bibinfo{journal}{\emph{ACM Trans. Graph.}} \bibinfo{volume}{29},
  \bibinfo{number}{4}, Article \bibinfo{articleno}{107} (\bibinfo{date}{July}
  \bibinfo{year}{2010}), \bibinfo{numpages}{8}~pages.
\newblock
\showISSN{0730-0301}
\urldef\tempurl%
\url{https://doi.org/10.1145/1778765.1778844}
\showDOI{\tempurl}


\bibitem[\protect\citeauthoryear{Wang, Liu, Zhu, Liu, Tao, Kautz, and
  Catanzaro}{Wang et~al\mbox{.}}{2018a}]%
        {wang2018vid2vid}
\bibfield{author}{\bibinfo{person}{Ting-Chun Wang}, \bibinfo{person}{Ming-Yu
  Liu}, \bibinfo{person}{Jun-Yan Zhu}, \bibinfo{person}{Guilin Liu},
  \bibinfo{person}{Andrew Tao}, \bibinfo{person}{Jan Kautz}, {and}
  \bibinfo{person}{Bryan Catanzaro}.} \bibinfo{year}{2018}\natexlab{a}.
\newblock \showarticletitle{Video-to-Video Synthesis}. In
  \bibinfo{booktitle}{\emph{Advances in Neural Information Processing Systems
  (NeurIPS)}}. \bibinfo{pages}{1152--–1164}.
\newblock


\bibitem[\protect\citeauthoryear{Wang, Liu, Zhu, Tao, Kautz, and
  Catanzaro}{Wang et~al\mbox{.}}{2018b}]%
        {WangLZTKC2018}
\bibfield{author}{\bibinfo{person}{Ting-Chun Wang}, \bibinfo{person}{Ming-Yu
  Liu}, \bibinfo{person}{Jun-Yan Zhu}, \bibinfo{person}{Andrew Tao},
  \bibinfo{person}{Jan Kautz}, {and} \bibinfo{person}{Bryan Catanzaro}.}
  \bibinfo{year}{2018}\natexlab{b}.
\newblock \showarticletitle{High-Resolution Image Synthesis and Semantic
  Manipulation with Conditional {GANs}}. In \bibinfo{booktitle}{\emph{CVPR}}.
\newblock


\bibitem[\protect\citeauthoryear{Wang, Bagautdinov, Lombardi, Simon, Saragih,
  Hodgins, and ZollhÃ¶fer}{Wang et~al\mbox{.}}{2020}]%
        {wang2020learning}
\bibfield{author}{\bibinfo{person}{Ziyan Wang}, \bibinfo{person}{Timur
  Bagautdinov}, \bibinfo{person}{Stephen Lombardi}, \bibinfo{person}{Tomas
  Simon}, \bibinfo{person}{Jason Saragih}, \bibinfo{person}{Jessica Hodgins},
  {and} \bibinfo{person}{Michael ZollhÃ¶fer}.}
  \bibinfo{year}{2020}\natexlab{}.
\newblock \bibinfo{title}{Learning Compositional Radiance Fields of Dynamic
  Human Heads}.
\newblock
\newblock
\showeprint[arxiv]{2012.09955}~[cs.CV]


\bibitem[\protect\citeauthoryear{Xian, Huang, Kopf, and Kim}{Xian
  et~al\mbox{.}}{2020}]%
        {Xian20arxiv_stnif}
\bibfield{author}{\bibinfo{person}{Wenqi Xian}, \bibinfo{person}{Jia-Bin
  Huang}, \bibinfo{person}{Johannes Kopf}, {and} \bibinfo{person}{Changil
  Kim}.} \bibinfo{year}{2020}\natexlab{}.
\newblock \showarticletitle{Space-time Neural Irradiance Fields for
  Free-Viewpoint Video}.
\newblock \bibinfo{journal}{\emph{https://arxiv.org/abs/2011.12950}}
  (\bibinfo{year}{2020}).
\newblock


\bibitem[\protect\citeauthoryear{Xu, Liu, Stoll, Tompkin, Bharaj, Dai, Seidel,
  Kautz, and Theobalt}{Xu et~al\mbox{.}}{2011}]%
        {Xu:SIGGRPAH:2011}
\bibfield{author}{\bibinfo{person}{Feng Xu}, \bibinfo{person}{Yebin Liu},
  \bibinfo{person}{Carsten Stoll}, \bibinfo{person}{James Tompkin},
  \bibinfo{person}{Gaurav Bharaj}, \bibinfo{person}{Qionghai Dai},
  \bibinfo{person}{Hans-Peter Seidel}, \bibinfo{person}{Jan Kautz}, {and}
  \bibinfo{person}{Christian Theobalt}.} \bibinfo{year}{2011}\natexlab{}.
\newblock \showarticletitle{Video-based Characters: Creating New Human
  Performances from a Multi-view Video Database}. In
  \bibinfo{booktitle}{\emph{ACM SIGGRAPH 2011 Papers}} (Vancouver, British
  Columbia, Canada) \emph{(\bibinfo{series}{SIGGRAPH '11})}.
  \bibinfo{publisher}{ACM}, \bibinfo{address}{New York, NY, USA}, Article
  \bibinfo{articleno}{32}, \bibinfo{numpages}{10}~pages.
\newblock
\showISBNx{978-1-4503-0943-1}
\urldef\tempurl%
\url{https://doi.org/10.1145/1964921.1964927}
\showDOI{\tempurl}


\bibitem[\protect\citeauthoryear{Xu, Umentani, Chao, Mao, Jin, and Tong}{Xu
  et~al\mbox{.}}{2014}]%
        {xuww14}
\bibfield{author}{\bibinfo{person}{Weiwei Xu}, \bibinfo{person}{Nobuyuki
  Umentani}, \bibinfo{person}{Qianwen Chao}, \bibinfo{person}{Jie Mao},
  \bibinfo{person}{Xiaogang Jin}, {and} \bibinfo{person}{Xin Tong}.}
  \bibinfo{year}{2014}\natexlab{}.
\newblock \showarticletitle{Sensitivity-Optimized Rigging for Example-Based
  Real-Time Clothing Synthesis}.
\newblock \bibinfo{journal}{\emph{ACM Trans. Graph.}} \bibinfo{volume}{33},
  \bibinfo{number}{4}, Article \bibinfo{articleno}{107} (\bibinfo{date}{July}
  \bibinfo{year}{2014}), \bibinfo{numpages}{11}~pages.
\newblock
\showISSN{0730-0301}
\urldef\tempurl%
\url{https://doi.org/10.1145/2601097.2601136}
\showDOI{\tempurl}


\bibitem[\protect\citeauthoryear{Yoon, Liu, Golyanik, Sarkar, Park, and
  Theobalt}{Yoon et~al\mbox{.}}{2020}]%
        {yoon2020poseguided}
\bibfield{author}{\bibinfo{person}{Jae~Shin Yoon}, \bibinfo{person}{Lingjie
  Liu}, \bibinfo{person}{Vladislav Golyanik}, \bibinfo{person}{Kripasindhu
  Sarkar}, \bibinfo{person}{Hyun~Soo Park}, {and} \bibinfo{person}{Christian
  Theobalt}.} \bibinfo{year}{2020}\natexlab{}.
\newblock \bibinfo{title}{Pose-Guided Human Animation from a Single Image in
  the Wild}.
\newblock
\newblock
\showeprint[arxiv]{2012.03796}~[cs.CV]


\bibitem[\protect\citeauthoryear{Zhang, Riegler, Snavely, and Koltun}{Zhang
  et~al\mbox{.}}{2020a}]%
        {Zhang20arxiv_nerf++}
\bibfield{author}{\bibinfo{person}{Kai Zhang}, \bibinfo{person}{Gernot
  Riegler}, \bibinfo{person}{Noah Snavely}, {and} \bibinfo{person}{Vladlen
  Koltun}.} \bibinfo{year}{2020}\natexlab{a}.
\newblock \showarticletitle{{NERF++}: Analyzing and Improving Neural Radiance
  Fields}.
\newblock \bibinfo{journal}{\emph{https://arxiv.org/abs/2010.07492}}
  (\bibinfo{year}{2020}).
\newblock


\bibitem[\protect\citeauthoryear{Zhang, Wang, Ceylan, and Mitra}{Zhang
  et~al\mbox{.}}{2020b}]%
        {zhangm20}
\bibfield{author}{\bibinfo{person}{Meng Zhang}, \bibinfo{person}{Tuanfeng
  Wang}, \bibinfo{person}{Duygu Ceylan}, {and} \bibinfo{person}{Niloy~J.
  Mitra}.} \bibinfo{year}{2020}\natexlab{b}.
\newblock \bibinfo{title}{Deep Detail Enhancement for Any Garment}.
\newblock
\newblock
\showeprint[arxiv]{2008.04367}~[cs.GR]


\bibitem[\protect\citeauthoryear{Zhi, Lassner, Tung, Stoll, Narasimhan, and
  Vo}{Zhi et~al\mbox{.}}{2020}]%
        {zhi2020texmesh}
\bibfield{author}{\bibinfo{person}{Tiancheng Zhi}, \bibinfo{person}{Christoph
  Lassner}, \bibinfo{person}{Tony Tung}, \bibinfo{person}{Carsten Stoll},
  \bibinfo{person}{Srinivasa~G. Narasimhan}, {and} \bibinfo{person}{Minh Vo}.}
  \bibinfo{year}{2020}\natexlab{}.
\newblock \bibinfo{title}{TexMesh: Reconstructing Detailed Human Texture and
  Geometry from RGB-D Video}.
\newblock
\newblock
\showeprint[arxiv]{2008.00158}~[cs.CV]


\bibitem[\protect\citeauthoryear{Zitnick, Kang, Uyttendaele, Winder, and
  Szeliski}{Zitnick et~al\mbox{.}}{2004}]%
        {zitnick04}
\bibfield{author}{\bibinfo{person}{C~Lawrence Zitnick},
  \bibinfo{person}{Sing~Bing Kang}, \bibinfo{person}{Matthew Uyttendaele},
  \bibinfo{person}{Simon Winder}, {and} \bibinfo{person}{Richard Szeliski}.}
  \bibinfo{year}{2004}\natexlab{}.
\newblock \showarticletitle{High-quality video view interpolation using a
  layered representation}. In \bibinfo{booktitle}{\emph{ACM Transactions on
  Graphics (TOG)}}, Vol.~\bibinfo{volume}{23}. ACM, \bibinfo{pages}{600--608}.
\newblock


\bibitem[\protect\citeauthoryear{{Zurdo}, {Brito}, and {Otaduy}}{{Zurdo}
  et~al\mbox{.}}{2013}]%
        {zurdo13}
\bibfield{author}{\bibinfo{person}{J.~S. {Zurdo}}, \bibinfo{person}{J.~P.
  {Brito}}, {and} \bibinfo{person}{M.~A. {Otaduy}}.}
  \bibinfo{year}{2013}\natexlab{}.
\newblock \showarticletitle{Animating Wrinkles by Example on Non-Skinned
  Cloth}.
\newblock \bibinfo{journal}{\emph{IEEE Transactions on Visualization and
  Computer Graphics}} \bibinfo{volume}{19}, \bibinfo{number}{1}
  (\bibinfo{year}{2013}), \bibinfo{pages}{149--158}.
\newblock


\end{thebibliography}


\end{document}